\documentclass{article}

\usepackage{microtype}
\usepackage{graphicx}
\usepackage{booktabs} 

\usepackage{hyperref}

\usepackage{amsmath,amsfonts,bm}

\def\eqref#1{equation~\ref{#1}}

\def\1{\bm{1}}

\def\vone{{\bm{1}}}

\def\va{{\bm{a}}}

\def\vh{{\bm{h}}}

\def\vx{{\bm{x}}}
\def\vy{{\bm{y}}}

\def\mA{{\bm{A}}}

\def\mD{{\bm{D}}}

\def\mL{{\bm{L}}}

\def\mX{{\bm{X}}}

\DeclareMathAlphabet{\mathsfit}{\encodingdefault}{\sfdefault}{m}{sl}
\SetMathAlphabet{\mathsfit}{bold}{\encodingdefault}{\sfdefault}{bx}{n}

\newcommand{\E}{\mathbb{E}}

\newcommand{\R}{\mathbb{R}}

\usepackage{xcolor}
\usepackage{lipsum}

\definecolor{darkblue}{HTML}{1A254B}
\definecolor{lightblue}{HTML}{A7BED3}
\definecolor{blue}{HTML}{114083}
\definecolor{strongblue}{HTML}{016fff}
\definecolor{green}{HTML}{5fbb46}
\definecolor{darkgreen}{HTML}{013220}
\definecolor{forestgreen}{HTML}{228B22}
\definecolor{pink}{HTML}{F2545B}
\definecolor{red}{HTML}{A4243B}
\definecolor{orang}{HTML}{F28C28}
\definecolor{yellow}{HTML}{f99e01}

\usepackage{hyperref}
\usepackage{graphicx}
\usepackage{url}
\usepackage{booktabs} 
\usepackage{wrapfig}
\usepackage{array}
\usepackage[labelformat=simple]{subcaption}
\definecolor{cite_color}{HTML}{114083}
\definecolor{link_color}{HTML}{F28C28}  
\definecolor{url_color}{RGB}{153, 102,  0}
\definecolor{emp_color}{RGB}{0,0,255}
\hypersetup{
 colorlinks,
 citecolor=cite_color,
 linkcolor=link_color,
 urlcolor=link_color}
\usepackage{mathtools}
\usepackage{multirow}
\usepackage{amsthm}

\usepackage[accepted]{icml2024}

\usepackage{amsmath}
\usepackage{amssymb}
\usepackage{mathtools}
\usepackage{amsthm}

\usepackage[capitalize,noabbrev]{cleveref}

\theoremstyle{plain}

\theoremstyle{definition}

\theoremstyle{remark}

\usepackage[textsize=tiny]{todonotes}

\icmltitlerunning{Harmonic Prior Flow Matching}
\begin{document}

\twocolumn[
\icmltitle{Harmonic Self-Conditioned Flow Matching \\for joint Multi-Ligand Docking and Binding Site Design}


\icmlsetsymbol{equal}{*}

\begin{icmlauthorlist}
\icmlauthor{Hannes Stark}{yyy}
\icmlauthor{Bowen Jing}{yyy}
\icmlauthor{Regina Barzilay}{yyy}
\icmlauthor{Tommi Jaakkola}{yyy}
\end{icmlauthorlist}

\icmlaffiliation{yyy}{CSAIL, Massachusetts Institute of Technology}

\icmlcorrespondingauthor{Hannes Stark}{hstark@mit.edu}

\icmlkeywords{Machine Learning, ICML}

\vskip 0.3in
]



\printAffiliationsAndNotice{}  
\begin{abstract}
    A significant amount of protein function requires binding small molecules, including enzymatic catalysis. As such, designing binding pockets for small molecules has several impactful applications ranging from drug synthesis to energy storage. Towards this goal, we first develop \textsc{HarmonicFlow}, an improved generative process over 3D protein-ligand binding structures based on our self-conditioned flow matching objective. \textsc{FlowSite} extends this flow model to jointly generate a protein pocket's discrete residue types and the molecule's binding 3D structure. We show that \textsc{HarmonicFlow} improves upon state-of-the-art generative processes for docking in simplicity, generality, and average sample quality in pocket-level docking. Enabled by this structure modeling, \textsc{FlowSite} designs binding sites substantially better than baseline approaches.
\end{abstract}

\section{Introduction}
Designing proteins that can bind small molecules has many applications, ranging from drug synthesis to energy storage or gene editing. Indeed, a key part of any protein's function derives from its ability to bind and interact with other molecular species. For example, we may design proteins that act as antidotes that sequester toxins or design enzymes that enable chemical reactions through catalysis, which plays a major role in most biological processes. We develop \textsc{FlowSite} to address this design challenge by building on recent advances in deep learning (DL) based protein design \citep{ProteinMPNN} and protein-molecule docking \citep{corso2023diffdock}.

\begin{figure}
  \centering
    \includegraphics[width=0.8\columnwidth]{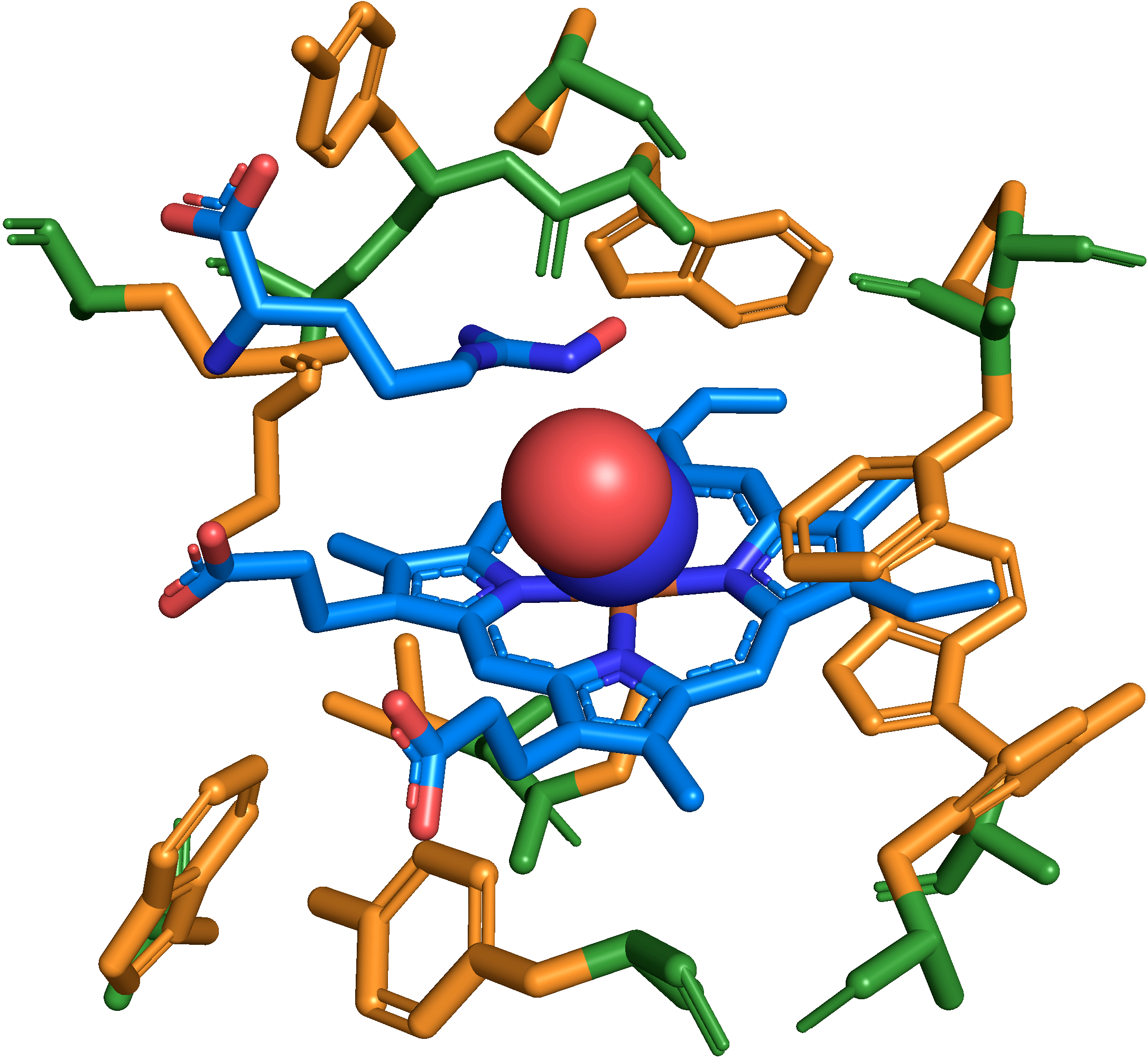}
  \caption{\textbf{Binding site design.} Given the backbone (\textcolor{forestgreen}{green}) and multi-ligand without structure, \textsc{FlowSite} generates residue types and structure (\textcolor{orange}{orange}) to bind the multi-ligand and its jointly generated structure (\textcolor{blue}{blue}). The majority of the pocket is omitted for visibility.} \label{fig:multi_ligand_task_explanation}
  \vspace{-0.0cm}
\end{figure}

Specifically, we aim to design protein pockets to bind a certain small molecule (called ligand). We assume that we are given a protein pocket via the 3D backbone atom locations of its residues as well as the 2D chemical graph of the ligand. We do not assume any knowledge of the 3D structure or the binding pose of the ligand. Based on this information, our goal is to predict the amino acid identities for the given backbone locations (see Figure \ref{fig:multi_ligand_task_explanation}). We also consider the more challenging task of designing pockets that simultaneously bind multiple molecules and ions (which we call multi-ligand). Such multi-ligand binding proteins are important, for example, in enzyme design, where the ligands correspond to reactants.

This task has not been addressed by deep learning yet. While deep learning has been successful in designing proteins that can bind to other proteins \citep{Watson2023rfdiffusion}, designing (multi-)ligand binders is different and arguably harder in various aspects. For example, no evolutionary information is directly available, unlike when modeling interactions between amino acids only. The existing approaches, such as designing 6 drug binding proteins \citet{polizzi2020structuralunit} or a single enzyme \citet{Yeh2023}, build on expert knowledge and require manual steps. Therefore, we develop \textsc{FlowSite} as a more general and automated approach and the first deep learning solution for designing pockets that bind small molecules.

\begin{figure*}[t]
    \vspace{-0.0cm}
    \centering
    \includegraphics[width=1.0\textwidth]{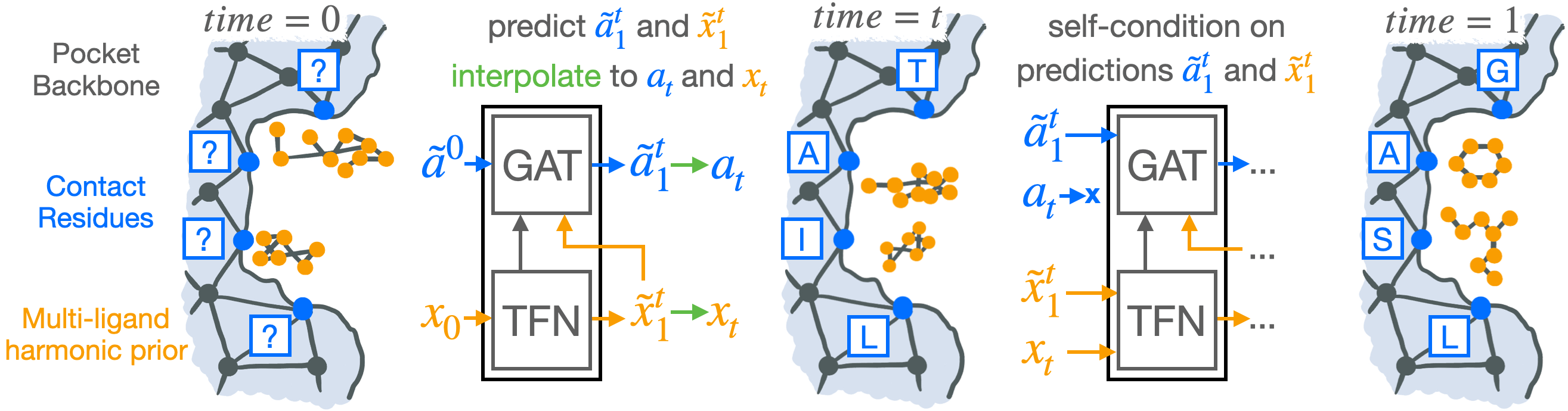}
    \vspace{-0.0cm}
  \caption{\textbf{Overview of FlowSite.} The generative process starts from a protein pocket's backbone atoms, initial residue types \textcolor{strongblue}{$\Tilde{\va}^0$}, and initial ligand positions \textcolor{yellow}{${\vx}_0$}. Our joint discrete-continuous self-conditioned flow updates them to \textcolor{strongblue}{${\va}_t$}, \textcolor{yellow}{${\vx}_t$} by \textcolor{green}{following its vector field} defined by the model outputs \textcolor{strongblue}{$\Tilde{\va}_1^t$}, \textcolor{yellow}{$\Tilde{\vx}_1^t$}. This integration is repeated until reaching $time=1$ with the produced sample \textcolor{strongblue}{${\va}_1$}, \textcolor{yellow}{${\vx}_1$}.} \label{fig:figure1}
  \vspace{-0.0cm}
\end{figure*}

\textsc{FlowSite} jointly generates discrete (residue identities) and continuous (ligand pose) variables. Our flow matching training criterion guides the model to learn a self-conditioned flow that jointly generates the contact residues and the (multi-)ligand 3D binding pose structures. To achieve this, we first develop \textsc{HarmonicFlow} as a suitable generative process for 3D poses of (multi-)ligands. \textsc{FlowSite} extends this process to residue types. Starting from initial residue types and ligand atom locations sampled from a harmonic prior \textsc{FlowSite} updates them by iteratively following the learned vector field, as illustrated in Figure \ref{fig:figure1}. 

The \textsc{HarmonicFlow} component of \textsc{FlowSite} performs the task known as docking, i.e., it realizes the 3D binding pose of the multi-ligand. As a method, it is remarkably simple in comparison to existing generative processes for docking, including the state-of-the-art product-space diffusion process of \textsc{DiffDock} \citep{corso2023diffdock} that operates on ligand's center of mass, orientation, and torsion angles. \textsc{HarmonicFlow} simply updates the cartesian coordinates of the atoms, yet manages to produce chemically plausible molecular structures without restricting ligand flexibility to torsions. Moreover, \textsc{HarmonicFlow} outperforms product-space diffusion in average sample quality on multiple pocket-level docking tasks on PDBBind. 

Having established \textsc{HarmonicFlow} as an improved generative process over ligand positions, we extend it to include discrete residue types to obtain \textsc{FlowSite}. We also adopt an additional "fake-ligand" data augmentation step where side chains are treated as ligands in order to realize additional training cases. Altogether, \textsc{FlowSite} is able to recover 47.0\% of binding site amino acids compared to 39.4\% of a baseline approach. This nearly closes the gap to an oracle method (51.4\% recovery) with access to the ground truth 3D structure/pose of the ligand. Next to technical innovations (self-conditioned flow matching or equivariant refinement TFN layers)      our main contributions are: 
\begin{enumerate}
\itemsep0.1cm 
    \item \textsc{FlowSite} as the first deep learning solution to design binding sites for small molecules without prior knowledge of the molecule structure.
    \item The \textsc{FlowSite} framework as a simple approach to jointly generate discrete and continuous data.
    \item \textsc{HarmonicFlow} which improves upon the state-of-the-art generative process for generating 3D ligand binding structures in average sample quality, simplicity, and applicability/generality.

\end{enumerate}

\section{Related Work}\label{sec:related_work}
\textbf{Deep learning for Docking.} Designing binding sites with high affinity for a ligand requires reasoning about the binding free energy, which is deeply interlinked with modeling ligand binding 3D structures. This task of molecular docking has recently been tackled with deep-learning approaches \citep{stärk2022equibind, Lu2022TankBind, zhang2023e3bind} including generative models \citep{corso2023diffdock, qiao2023statespecific}. These generative methods are based on diffusion models, building on \textsc{DiffDock} \citep{corso2023diffdock}, which combines diffusion processes over the ligand's torsion angles and position with respect to the protein. For the task of multi-ligand docking, no deep learning solutions exist yet, and we provide the first with \textsc{HarmonicFlow}. We refer to Appendix \ref{appx:related_wok} for additional important related work on this and the following topics.

\textbf{Protein Design.} A significant technical challenge for protein design is jointly modeling structure and sequence. Inverse folding approaches \citep{ProteinMPNN, gao2023pifold, yi2023graph, Hsu2022esmIF, gao2023knowledgedesign} attempt this by designing new sequences given a protein structure. Existing \emph{ligand aware} inverse folding methods such as Carbonara \citep{lucienkrapp2023carbonara} and LigandMPNN \citep{dauparas2023ligandmpnn} are limited in their applicability for small molecule binding site design since they assume the bound ligand structure to be provided as input instead of solving the hard problem of docking jointly with designing the backbone residues as achieved by \textsc{FlowSite}. 

The same limitation applies to the classical energy function and search algorithm based \textsc{PocketOptimizer} \citep{Noske2023-th} and the sequence-structure co-design framework \textsc{FAIR} \citep{zhang2023fullatom}, which both require the bound complex as input. \textsc{FAIR} is further distinct from \textsc{FlowSite} in that it uses all residue types of a protein as input except for those in contact with the ligand, which simplifies recovering the amino acids of known binders to infilling missing residues based on sequence similarity.

\textbf{Flow Matching.} This recent generative modeling paradigm \citep{lipman2022flow, albergo2022building, albergo2023stochastic} generalizes diffusion models \citep{ho2020denoising, song2021score} in a simpler framework. Flow matching admits more design flexibility and multiple works \citep{tong2023improving, pooladian2023multisample} showed how it enables learning flows between arbitrary start and end distributions in a simulation-free manner. It is easily extended to data on manifolds \citep{chen2023riemannian} and comes with straighter paths that enable faster integration. 

Other applications of flow matching to biomolecular problems include generating Boltzmann distributions of small molecules \citet{klein2023equivariant}, protein structure generation \citep{yim2023fast, bose2023se3stochastic} and small molecule generation \citep{song2023equivariant}. We explain flow matching in Section \ref{sec:harmonic_flow}.


\section{Method} \label{sec:method}

Our goal is to design binding pockets for a ligand where we assume the inputs to be the ligand's 2D chemical graph and the backbone coordinates of the pocket's residues. In this section, we lay out how \textsc{FlowSite} achieves this by first explaining our \textsc{HarmonicFlow} generative process for docking in \ref{sec:harmonic_flow} before covering how \textsc{FlowSite} extends it to include discrete residue types in \ref{sec:method_flowsite} and concluding with our model architecture in \ref{sec:method_architecture}. 

\textbf{Overview and definitions.} As visualized in Figure \ref{fig:figure1}, \textsc{FlowSite} jointly updates discrete residue types and continuous ligand positions. The inputs are a protein pocket's backbone atoms $\vy \in \R^{L \times 4 \times 3}$ for $L$ residues with 4 atoms each and the chemical graph of a (multi-)ligand. Based on the ligand connectivity, its initial coordinates $\vx \in \R^{n \times 3}$ are sampled from a harmonic prior, and we initialize residue types $\va \in \{1, \dots, 20\}^L$ with an initial token (we drop the chemical information of the ligands in our notation for brevity). 

Given this at time $t=0$, the flow model $v_\theta$ with learned parameters $\theta$ iteratively updates residue types and ligand coordinates by integrating the ODE it defines. These integration steps are repeated from time $t=0$ to time $t=1$ to obtain the final generated binding pocket designs.

\subsection{HarmonicFlow Structure Generation}\label{sec:harmonic_flow}
We first lay out \textsc{HarmonicFlow} for pure structure generation without residue type estimation. Our notation drops $v_\theta$'s conditioning on the pocket $\vy$ and residue estimates $\va$ in this subsection (see the Architecture Section \ref{sec:method_architecture} for how $\vy$ is included). Simply put, \textsc{HarmonicFlow} is flow matching with a harmonic prior, self-conditioning, and $\vx_1$ prediction (our refinement TFN layers in Section \ref{sec:method_architecture} are also important for performance). In more detail:

\label{par:conditional-flow-matching}
\textbf{Conditional Flow Matching.} Given the data distribution $p_1$ of bound ligand structures and any easy-to-sample prior $p_0$ over $\R^{n\times 3}$, we wish to learn an ODE that pushes the prior forward to the data distribution when integrating it from time 0 to time 1. The ODE will be defined by a time-dependent vector field $v_\theta(\cdot, \cdot): \R^{n\times 3} \times [0,1] \mapsto \R^{n\times 3}$. Starting with a sample $\vx_0 \sim p_0(\vx_0)$ and following/integrating $v$ through time will produce a sample from the data distribution $p_1$.

To see how to train $v_\theta$, let us first assume access to a time-dependent vector field $u_t(\cdot)$ that would lead to an ODE that pushes from the prior $p_0$ to the data $p_1$ (it is not straightforward how to construct this $u_t$). This gives rise to a probability path $p_t$ by integrating $u_t$ until time $t$. If we could sample $\vx \sim p_t(\vx)$ we could train $v_\theta$ with the unconditional flow matching objective \citep{lipman2022flow}
\begin{equation}
    \mathcal{L}_{FM} = \E_{t \sim \mathcal{U}[0,1], \vx \sim p_t(\vx)}\lVert v_\theta(\vx,t) - u(\vx,t) \rVert^2.
\end{equation}

Among others, \citet{tong2023improving} show that to construct such a $u_t$ (that transports from prior $p_0$ to $p_1$), we can use samples from the data $\vx_1 \sim p_1(\vx_1) $ and prior $\vx_0 \sim p_0(\vx_0) $ and define $u_t$ via 

\begin{multline}
    u_t(\vx) = \E_{\vx_1 \sim p_1(\vx_1), \vx_0 \sim p_0(\vx_0)} \\ \frac{u_t(\vx|\vx_0,\vx_1) p_t(\vx|\vx_0,\vx_1)}{p_t(\vx)}
\end{multline}

where we can choose easy-to-sample conditional flows $p_t(\cdot|\cdot,\cdot)$ that give rise to simple conditional vector fields $u_t(\cdot|\cdot,\cdot)$. 
We still cannot efficiently compute this $u_t(\vx)$ and use it in $\mathcal{L}_{FM}$ because we do not know $p_t(\vx)$, but there is no need to: it is equivalent to instead train with the following conditional flow matching loss since $\nabla_\theta\mathcal{L}_{FM} = \nabla_\theta\mathcal{L}_{CFM}$.

\begin{multline}
    \mathcal{L}_{CFM} = \E_{t \sim \mathcal{U}[0,1], \vx_1 \sim p_1(\vx_1), \vx_0 \sim p_0(\vx_0), x \sim p_t(\vx|\vx_0,\vx_1)} \\ \lVert v_\theta(\vx,t) - u_t(\vx|\vx_0,\vx_1) \rVert^2.
\end{multline}

Our simple choice of conditional probability path is $p_t(\vx|\vx_0,\vx_1) = \mathcal{N}(\vx | t\vx_1 + (1-t)\vx_0, \sigma^2)$, which gives rise to the conditional vector field $u_t(\vx|\vx_0,\vx_1) = \vx_1 - \vx_0$. Notably, we find it helpful to parameterize $v_\theta$ to predict $\vx_1$ instead of $(\vx_1 - \vx_0)$. 

{\color{strongblue} \textit{Training}} with the conditional flow matching loss then boils down to 1) Sample data $\vx_1 \sim p_1(\vx_1) $ and prior $\vx_0 \sim p_0(\vx_0) $. 2) Interpolate between between the points. 3) Add noise to the interpolation to obtain $x$. 4) Evaluate and minimize $\mathcal{L}_{CFM} = \lVert v_\theta(\vx,t) - \vx_1 \rVert^2$ with it. {\color{strongblue} \textit{Inference}} is just as straightforward. We sample from the prior $\vx_0 \sim p_0(\vx_0)$ and integrate from $t=0$ to $t=1$ with an arbitrary ODE solver. We use an Euler solver, i.e., we iteratively predict $\vx_1$ as $\Tilde{\vx}_1 = v_\theta(\vx_t,t)$, and then calculate the step size scaled velocity estimate from it and add it to the current point $\vx_{t + \Delta t} = \vx_t + \Delta t(\Tilde{\vx}_1 - \vx_0)$. Training and inference algorithms are in Appendix \ref{appx:method_algorithms}.

\begin{figure}
  \centering
    \includegraphics[width=\columnwidth]{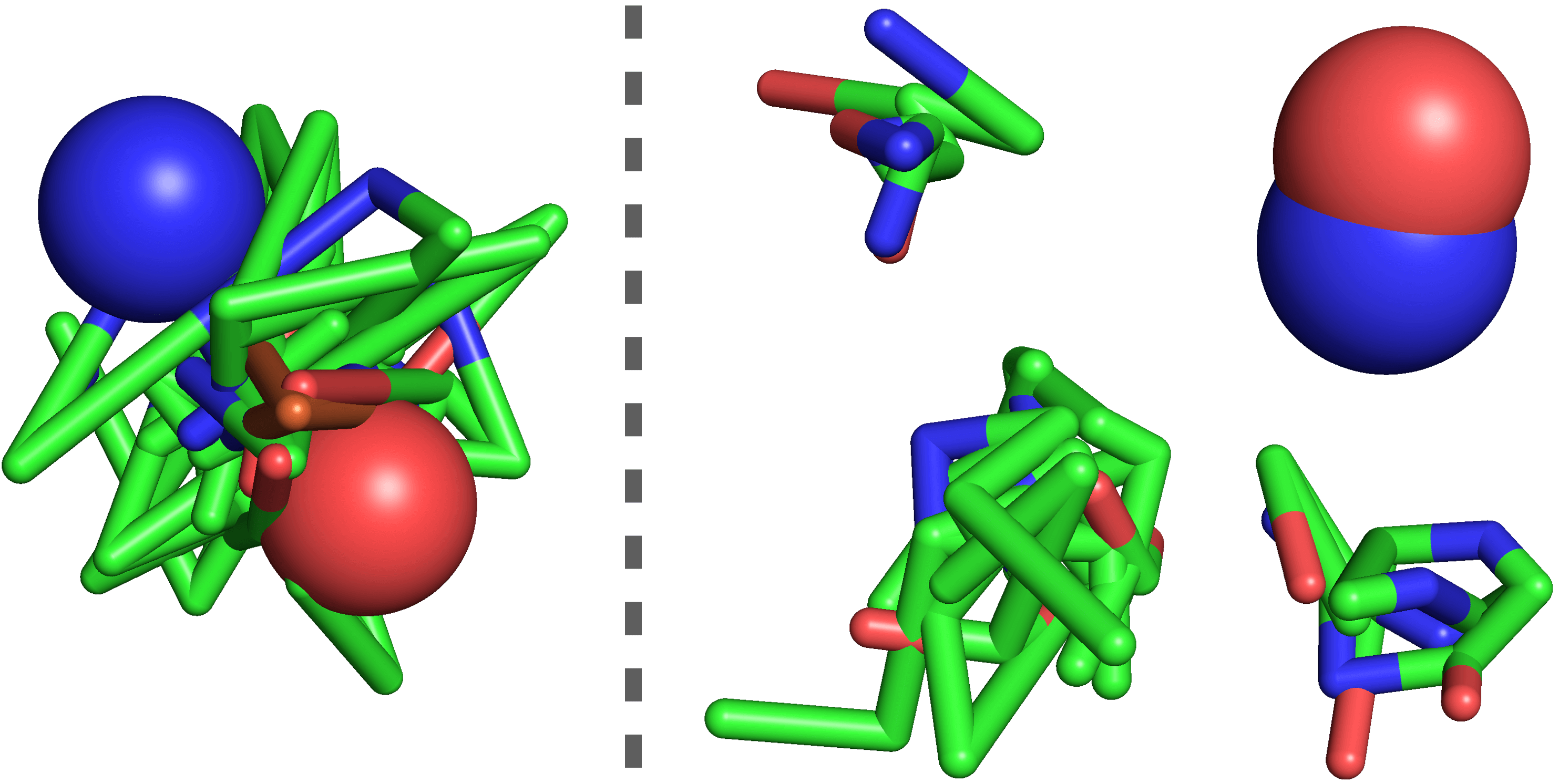}
  \caption{\textbf{Harmonic Prior.} Initial positions for the same single multi-ligand from an isotropic Gaussian (\textit{left}) and from a harmonic prior (\textit{right}).  (Bound structure for this multi-ligand is in Figure \ref{fig:multi_ligand_task_explanation}).} \label{fig:harmonic_prior}
  \vspace{-0.0cm}
\end{figure}
\textbf{Harmonic Prior.} Any prior can be used for $p_0$ in the flow matching framework. We choose a harmonic prior as in EigenFold \citep{jing2023eigenfold} that samples atoms to be close to each other if they are connected by a bond. Potentially, this inductive bias is especially helpful when dealing with multiple molecules and ions since atoms of different molecules are already spatially separated at $t=0$ as visualized in Figure \ref{fig:harmonic_prior}. 

This prior is constructed based on the ligand's covalent bonds that define a graph with adjacency matrix $\mA$ from which we can construct the graph Laplacian $\mL = \mD - \mA$ where $\mD$ is the degree matrix. The harmonic prior is then $p_0(\vx_0) \propto exp(-\frac{1}{2}\vx_0^T\mL\vx_0)$ which can be sampled as a transformed gaussian.

\label{method:structure-self-conditioning}
\textbf{Structure Self-conditioning.} With this, we aim to bring AlphaFold2's \citep{jumper2021highly} successful recycling strategy to flow models for structure generation. Recycling enables training a deeper structure predictor without additional memory cost by performing multiple forward passes while only computing gradients for the last. For flow matching, we achieve the same by adapting the discrete diffusion model self-conditioning approach of \citet{chen2023analog}, similar to self-conditioning in protein structure generation diffusion models \citep{yim2023se, Watson2023rfdiffusion}.

Instead of defining the vector field $v_\theta(\vx_t, t)$ as a function of $\vx_t$ and $t$ alone, we additionally condition it on the prediction $\Tilde{\vx}_1^t$ of the previous integration step and use $v_\theta(\vx_t, \Tilde{\vx}_1^t,t)$. At the beginning of \textcolor{strongblue}{\textit{inference}} the self-conditioning input is a sample from the harmonic prior $\Tilde{\vx}^{0}_1 \sim p_0(\Tilde{\vx}^{0}_1)$. In all following steps, it is the flow model's output (its prediction of $\vx_1$) of the previous step $\Tilde{\vx}^t_1 = v_\theta(\vx_{t-\Delta t}, \Tilde{\vx}_1^{t-\Delta t},t-\Delta t)$. To \textcolor{strongblue}{\textit{train}} this, in a random 50\% of the training steps, the self-conditioning input is a sample from the prior $\Tilde{\vx}^0_1$. In the other 50\%, we first generate a self-conditioning input $\Tilde{\vx}^{t+ \Delta t}_1 = v_\theta(\vx_t, \Tilde{\vx}^0_1,t)$, detach it from the gradient computation graph, and then use $v_\theta(\vx_t, \Tilde{\vx}^{t+ \Delta t}_1,t)$ for the loss computation. Algorithms \ref{alg:self_training} and \ref{alg:self_inference} show these training and inference procedures. 


\subsection{FlowSite Binding Site Design}\label{sec:method_flowsite}
\begin{figure}
  \centering
    \includegraphics[width=\columnwidth]{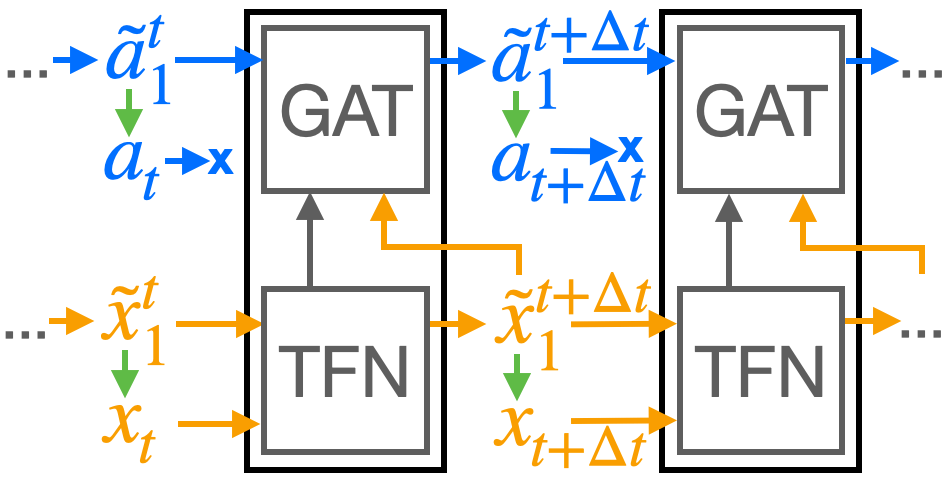}
  \vspace{-0.0cm}
  \caption{\textbf{FlowSite self-conditioned updates.} Residue type predictions \textcolor{strongblue}{$\Tilde{\va}_1^t$} from invariant GAT layers and position predictions \textcolor{yellow}{$\Tilde{\vx}_1^t$} from equivariant TFN layers are used as self-conditioning inputs and to \textcolor{green}{interpolate} to the updates \textcolor{strongblue}{${\va}_t$}, \textcolor{yellow}{${\vx}_t$}.}\label{fig:flowsite_updates}
  \vspace{-0.0cm}
\end{figure}

In the \textsc{FlowSite} binding site design framework, \textsc{HarmonicFlow} $\Tilde{\vx}^{t + \Delta t}_1 = v_\theta(\vx_t,\Tilde{\vx}^{t}_1,t)$ is augmented with an additional self-conditioned flow over the residue types to obtain $(\Tilde{\vx}^{t + \Delta t}_1, \Tilde{\va}^{t + \Delta t}_1) = v_\theta(\vx_t,\Tilde{\vx}^{t}_1, \va_t, \Tilde{\va}^{t}_1,t)$. The flow no longer produces $\Tilde{\vx}^{t + \Delta t}_1$ as an estimate of $\vx_1$ and then interpolates to $\vx_{t + \Delta t}$ but instead produces $(\Tilde{\vx}^{t + \Delta t}_1, \Tilde{\va}^{t + \Delta t}_1)$ from which we obtain the interpolation $(\vx_{t + \Delta t}, \va_{t + \Delta t})$ and use it for the next integration step (see Figure \ref{fig:flowsite_updates}). The start $\va_0, \Tilde{\va}^0_1$ are initialized as a mask token while the structures $\vx_0, \Tilde{\vx}^0_1$ are drawn from a harmonic prior.

This joint discrete-continuous data process is trained with the same self-conditioning strategy as in \hyperref[method:structure-self-conditioning]{structure self-conditioning}, but with the additional discrete self-conditioning input $\Tilde{\va}^1_1$ that is either a model output or a mask token. To the training loss we add the  cross-entropy $\mathcal{L}_{type}$ between $\va$ and $\Tilde{\va}^t_1$. In practice, we find that the $\va_1$ prediction $\Tilde{\va}^t_1$ already carries most information that is useful for predicting $a_1$ and we omit the interpolation $\va_t$ as model input to obtain the simpler $(\Tilde{\vx}^{t + \Delta t}_1, \Tilde{\va}^{t + \Delta t}_1) = v_\theta(\vx_t,\Tilde{\vx}^{t}_1, \Tilde{\va}^{t}_1,t)$. This formulation admits a direct interpretation as recycling \cite{jumper2021highly} and a clean joint discrete-continuous process without defining a discrete data interpolation.

\textbf{Fake Ligand Data Augmentation.} \label{sec:method-fake-ligand}
This strategy is based on the evidence of \citet{polizzi2020structuralunit} that a protein's sidechain-sidechain interactions are similar to sidechain-ligand interactions for tight binding. In our optional data augmentation, we train with 20\% of the samples having a \textit{"fake ligand"}. Given a protein, we construct a fake ligand as the atoms of a randomly selected residue that has at least 4 other residues within 4\AA{} heavy atom distance. Additionally, we modify the protein by removing the residue that was chosen as the fake ligand and the residues that are within 7 positions next to that residue in the protein chain (Figure \ref{fig:fake_ligand}). This data augmentation was also employed in concurrent work \citep{dauparas2023ligandmpnn, corso2024deep}.

\subsection{Architecture} \label{sec:method_architecture}

Here, we provide an overview of the \textsc{FlowSite} architecture (visualized in Appendix Figure \ref{fig:architecture}) that outputs ligand positions $\Tilde{\vx}_1$ and uses them for a residue type prediction $\Tilde{\va}_1$. The structure prediction is produced by a stack of our SE(3)-equivariant refinement TFN layers that are crucial for the performance of \textsc{HarmonicFlow}'s structure generation. This is followed by invariant layers to predict the invariant residue types, which we found to perform better than the more expressive equivariant layers that seem well suited for structure prediction but not for recovering residue types. The precise architecture definition is in Appendix \ref{appx:method_architecture} and an architecture visualization in Figure \ref{fig:architecture}.

\textbf{Radius Graph Representation.} We represent the (multi-)ligand and the protein as graphs where nodes are connected based on their distances. Each protein residue and each ligand atom is a node. These are connected by protein-to-protein edges, ligand-to-ligand edges, and edges between ligand and protein. While only a single node is assigned to each residue, they contain information about all backbone atom positions (N, Ca, C, O).

\textbf{Equivariant refinement TFN layers.} \label{par:refinement-tfn}
Based on Tensor Field Networks (TFN) \citep{e3nn}, these layers are a simple yet effective tweak from previous use cases of message passing TFNs \citep{jing2022torsional, corso2023diffdock}, where we instead update and refine ligand coordinates with each layer akin to EGNNs \citep{hoogeboom2022equivariant}. See Appendix \ref{appx:method_architecture} for more details.

The $k$-th refinement TFN layer takes as input the protein positions $\vy$, current ligand positions $\vx_t$, and features $\vh^{k-1}$ (with $\vh^{0}$ being zeros for the ligand and vectors between N, Ca, C, O for the protein). We construct equivariant messages for each edge via a tensor-product of neighboring nodes' invariant and equivariant features. The messages include the \hyperref[method:structure-self-conditioning]{structure self-conditioning} information by using the interatomic distances of the self-conditioning input $\vx^{t}_1$ to parameterize the tensor products. We sum the messages to obtain new node features $\vh^{k+1}$ and use them as input to an O(3) equivariant linear layer to predict intermediate refined ligand coordinates $\hat{\vx}_1^k$. Before passing $\hat{\vx}_1^k$ to the next refinement TFN layer, we detach them from the gradient computation graph for the non-differentiable radius graph building of the next layer. 

After a stack of $K$ TFN refinement layers, the positions $\hat{\vx}_1^K$ are used as final prediction $\Tilde{\vx}^{t + \Delta t}_1$. While $\Tilde{\vx}^{t + \Delta t}_1$ is supervised with the conditional flow matching loss $\mathcal{L}_{CFM} = \lVert \Tilde{\vx}^{t + \Delta t}_1 - \vx_1 \rVert^2$ the intermediate positions $\hat{\vx}_1^k$ contribute to an additional refinement loss $\mathcal{L}_{refine} = \sum^{K-1}_{k=1}\lVert \hat{\vx}_1^k - \vx_1 \rVert^2$.

\textbf{Invariant Network.}
The inputs to this part of \textsc{FlowSite} are the TFN's ligand structure prediction $\Tilde{\vx}_1$, the protein structure $\vy$,  the invariant scalar features of the refinement TFN layers, and the self-conditioning input $\va^t_1$. From the protein structure, we construct on PiFold's \citep{gao2023pifold} distance-based invariant edge features and node features that encode the geometry of the backbone. For the edges between protein and ligand, we construct features that encode the distances from a ligand atom to all 4 backbone atoms of a connected residue. 

These are processed by a stack of graph attention layers that update ligand and protein node features as well as edge features for each type of edge (ligand-to-ligand, protein-to-protein, and between the molecules). For each edge, the convolutional layers first predict attention weights from the edge features and the features of the nodes they connect. We then update a node's features by summing messages from each incoming edge weighted by the attention weights. Then, we update an edge's features based on its nodes' new features. A precise definition is in Appendix \ref{appx:method_architecture}. From the residue features after a stack of these convolutions, we predict new residue types $\va_{t+\Delta t}$ together with side chain torsion angles $\alpha$. We use those in an auxiliary loss $\mathcal{L}_{torsion}$ defined as in AlphaFold2's Appendix 1.9.1 \citep{jumper2021highly}. Thus, the complete loss for \textsc{FlowSite} is a weighted sum of $\mathcal{L}_{CFM}, \mathcal{L}_{refine}, \mathcal{L}_{type}$, and $\mathcal{L}_{torsion}$, while \textsc{HarmonicFlow} only uses $\mathcal{L}_{CFM}$ and $\mathcal{L}_{refine}$.

\section{Experiments} \label{sec:experiments}
We evaluate \textsc{FlowSite} with multiple datasets, splits, and settings. {Every reported number is averaged over 10 generated samples for each ligand.} Precise experimental details are in Appendix \ref{appx:experiment_details} and code to reproduce each experiment is at \url{https://github.com/HannesStark/FlowSite}. The main questions we seek to answer with the experiments are:

\begin{itemize}
    \item \textbf{1. Structure Generation:} How does \textsc{HarmonicFlow} compare with SOTA binding structure generative models? Does it work for \textit{multi-ligands}?
    \item \textbf{2. Binding Site recovery:} What is the improvement in recovering residue types of known binding sites with \textsc{FlowSite} compared with baseline approaches?
    \item \textbf{3. Ablations and Flow Matching Investigations:} How does flow matching behave for structure generation, and how much does each component help? (self-conditioning, ...)
\end{itemize}

\subsection{Datasets}
We use \textbf{PDBBind} version 2020 with 19k complexes to evaluate the structure generation capability of flow matching and the ability of \textsc{FlowSite} to design binders for a single connected ligand. We employ two dataset splits. The first is based on time, which has been heavily used in the DL community \citep{stärk2022equibind, corso2023diffdock}. The second is sequence-based with a maximum of 30\% chain-wise similarity between train, validation, and test data. \citet{buttenschoen2023posebusters} found DL docking methods to be significantly more challenged by sequence similarity splits.

For many binding pocket design tasks, it is required to bind multi-ligands. For example, when designing enzymes for multiple reactants. Such multi-ligands are present in \textbf{Binding MOAD}. We use its 41k complexes with a 30\% sequence similarity split carried out as described above. We construct our \textit{multi-ligands} as all molecules and ions that have atoms within 4\AA{} of each other. An example of an enzyme with all substrates in the pocket as multi-ligand is in Figure \ref{fig:multi_ligand_task_explanation}.

\subsection{Question 1: HarmonicFlow Structure Generation}\label{sec:experiments_structure_generation}
\begin{table*}[h]
\caption{\textbf{\textsc{HarmonicFlow} vs. \textsc{Product Space Diffusion}.} Comparison on PDBBind splits for docking into \textit{Distance-Pockets} (residues close to ligand) and \textit{Radius-Pockets} (residues within a radius of the pocket center). The columns "\%$<$2" show the fraction of predictions with an RMSD to the ground truth that is less than 2\AA{} (higher is better). "Med." is the median RMSD (lower is better).} \label{table:docking_diffdock}
\begin{center}
    \begin{tabular}{lcc|cc|cc|cc}
    \toprule
      & \multicolumn{4}{c}{Sequence Similarity Split} & \multicolumn{4}{c}{Time Split} \\ 
      \rule{0pt}{2ex}
      & \multicolumn{2}{c}{Distance-Pocket} & \multicolumn{2}{c}{Radius-Pocket} & \multicolumn{2}{c}{Distance-Pocket} & \multicolumn{2}{c}{Radius-Pocket} \\
    \rule{0pt}{2ex}  
        Method & \%$<$2 & Med. & \%$<$2 & Med. & \%$<$2 & Med. & \%$<$2 & Med. \\
    \midrule
    \textsc{Product Space Diffusion}  & 27.2 & 3.2 & 16.1 & 4.0 & 20.8 & 3.8 & 15.2 & 4.3  \\
    \textsc{HarmonicFlow} & 30.1 & 3.1 & 20.5 & 3.4 &  42.8 & 2.5 & 28.3 & 3.2 \\
    \bottomrule
    \end{tabular}
\end{center}
\vspace{-0.3cm}
\end{table*}
Here, we consider the \textsc{HarmonicFlow} component of \textsc{FlowSite} and investigate its binding structure generation capability. This is to find out whether \textsc{HarmonicFlow} is fit for binder design where good structure generation is necessary to takethe bound ligand structure into account. 

\textbf{Task Setup.} The architecture only contains refinement TFN layers, and there is no sequence prediction. The inputs are the (multi-)ligand's chemical graph and the protein pocket's backbone atoms and residue types (see Appendix Table \ref{table:blind_docking} for experiments without residue type inputs). From this, the binding structure of the (multi-)ligand has to be inferred. There is also no \hyperref[sec:method-fake-ligand]{\textit{fake ligand augmentation}}.

We test docking on the pocket level since that is the structure modeling capability required for the binding site design task (in Appendix \ref{appx:results}, we show preliminary results for docking to the whole protein). We define the binding pocket in two ways. {In the \textit{Distance-Pocket} definition, we calculate the distances of all ligand heavy atoms to the protein's alpha carbons, add Gaussian noise with $\sigma=0.5$ to them, and include a residue in the pocket if it has any noisy distance smaller than 14 \AA{}. The center of the pocket (where the prior $p_0$ is centered) is the center of mass of all residues within 8 \AA{} of noisy ligand distance. We additionally add Gaussian noise with  $\sigma=0.2$ to the pocket center. The motivation for the noisy distance cutoffs and pocket center is to alleviate distribution shifts during inference time and to prevent the models from inferring the ligand positions from the cutoff with which the pocket was constructed.}

{In \textit{Radius-Pockets}, we first obtain the center of mass of residues within 8 \AA{} of any ligand heavy atom. The pocket includes all residues with a noisy distance to the center of mass that is less than a specific radius. This radius is 7\AA{} plus the minimum of 5\AA{} and half of the ligand's diameter. The pocket center is obtained in the same way as for \textit{Distance-Pockets}.}


\textbf{Baseline.} We compare with the state-of-the-art product-space diffusion process of \textsc{DiffDock} \citep{corso2023diffdock} {which has recently also been proven successful for pocket level docking \citep{plainer2023diffdockpocket}}. \emph{Note that this is not the full \textsc{DiffDock} docking pipeline:} Both \textsc{HarmonicFlow} and \textsc{DiffDock}'s diffusion can generate multiple samples and, for the task of docking, a further discriminator (called confidence model in \textsc{DiffDock}) could be used to select the most likely poses. We only compare the 3D structure generative models and neither use language model residue embeddings. We train product-space diffusion with our pocket definitions using 5 of its default TFN layers followed by its pseudotorque and center-convolution (the TFN layers are identical to ours apart from our position updates). This uses their training parameters and the same 32 scalar and 8 vector features of our model. We use this deep learning method for comparison since (like \textsc{HarmonicFlow}) it is able to dock to protein structures without an all-atom representation (unlike traditional docking methods, e.g., Autodock VINA \citep{trott2010autodock}). This is required for the pocket design task where the sidechain atom locations are unknown.

\textbf{PDBBind docking results.}
{In Table \ref{table:docking_diffdock}, we find that our flow matching based \textsc{HarmonicFlow} outperforms product-space diffusion in average sample quality. The sampled conformations in Figure \ref{fig:visualizations} show that \textsc{HarmonicFlow} produces chemically plausible structures and well captures the physical constraints of interatomic interactions without the need to restrict conformational flexibility to torsion angles. No separate losses, rotation updates, or expensive torsion angle updates are required - \textsc{HarmonicFlow} is arguably a cleaner and simpler solution. Thus, it is a promising future direction to further explore \textsc{HarmonicFlow} for docking and other biomolecular structure generation tasks.}

\begin{table}
\vspace{-0.4cm}
\caption{\textbf{Multi-Ligand Docking.} Structure generation performance on Binding MOAD's \textit{multi-ligands}. "\%$<$2" means the fraction of predictions with an RMSD to the ground truth less than 2\AA{} (higher better). "Med." is the median RMSD (lower better).}
\label{table:docking_multi_ligand}
\begin{center}
    \begin{tabular}{lccc}
    \toprule
        Method & \%$<$2 & \%$<$5 & Med. \\
    \midrule
    \textsc{EigenFold Diffusion}     & 39.7 & 73.5 & 2.4 \\
    \textsc{HarmonicFlow}              & 44.4 & 75.0 & 2.2 \\
    \bottomrule
    \end{tabular}
\vspace{-0.4cm}
\end{center}
\end{table}

\textbf{Binding MOAD multi-ligand docking results.} For binding site design, it is often necessary to model multiple ligands and ions (e.g., reactants for an enzyme). We test this with Binding MOAD, which contains such multi-ligands. Since no deep learning solutions for multi-ligands exist yet and traditional docking methods would require side-chain atom locations, we compare with \textsc{EigenFold}'s \citep{jing2023eigenfold} Diffusion and provide qualitative evaluation in Appendix Figure \ref{fig:visualizations}. For \textsc{EigenFold Diffusion}, we use the same model as \textsc{HarmonicFlow} and predict $\vx_0$ (in what corresponds to $\vx_0$ prediction in diffusion models), which we found to work better. Table \ref{table:docking_multi_ligand} shows \textsc{HarmonicFlow} as viable for docking multi-ligands - thus, the first ML method for this task with important applications besides binding site design.  



\subsection{Question 2: FlowSite Binding Site Recovery}
\begin{table*}[h]
\caption{\textbf{Binding Site Recovery.} Comparison on PDBBind and Binding MOAD sequence similarity splits for recovering residues of binding sites. \textit{Recovery} is the percentage of correctly predicted residues, and  \textit{BLOSUM score} takes residue similarity into account. 2D ligand refers to a simple GNN encoding of the ligand's chemical graph as additional input. The \textsc{Ground Truth Pos} row has access to the, in practice, unknown ground truth 3D crystal structure of the ligand and protein.}
\label{table:pocket_recovery}
\vspace{-0.0cm}
\begin{center}
    \begin{tabular}{lcc|cc}
    \toprule
      & \multicolumn{2}{c}{Binding MOAD} & \multicolumn{2}{c}{PDBBind}  \\ 
      \rule{0pt}{2ex}
    \rule{0pt}{2ex}  
        Method &  \textit{BLOSUM score} & \textit{Recovery} &  \textit{BLOSUM score} & \textit{Recovery}   \\
    \midrule
    \textsc{ProteinMPNN} (no ligand)    & -- & --  & 40.3 & 36.3   \\
    \textsc{PiFold} (no ligand)         & 35.2 & 39.4  & 40.7 & 43.5   \\
    \textsc{PiFold} (2D ligand)         & 35.7 & 40.4  & 42.2 & 44.5   \\
    \textsc{Random Ligand Pos}          & 38.2 & 41.8  & 41.5 & 43.7   \\
    {\textsc{DiffDock-Pocket Pos}}          & {--} &{--}  & {42.6} & {45.0}   \\
    \midrule
    \textsc{FlowSite}                   & 44.3 & 47.0  & 47.6 & {49.5}   \\
    \midrule
    \textsc{Ground Truth Pos}           & 48.4 & 51.4  & 51.3 & 51.2   \\
    \bottomrule
    \end{tabular}
\vspace{-0.2cm}
\end{center}
\end{table*}


\textbf{Setup.} The input to \textsc{FlowSite} is the binding pocket/site specified by its backbone and the chemical identity of the ligand (without its 3D structure). With the pocket, in practice, chosen by the user and well known, we use the \textit{Distance-Pocket} definition here. 

\textbf{Metrics.} We use two metrics, sequence recovery and our \textit{BLOSUM score}. Sequence recovery is the percentage of generated residue types of the contact residues (those with a heavy atom within 4\AA{} of the ligand) that are the same as in the original binding site. This metric only rewards exact matches, {it cannot account for multiple correct solutions}, and there is no notion of amino acid similarity encoded in it. Having lower penalties for mismatches of very similar amino acids would be more meaningful. To address this, we propose \textit{BLOSUM score}, which takes evolutionary and physicochemical similarity into account. A precise definition is in Appendix \ref{appx:blosum_score}.

We note that metrics that correlate with binding affinity, such as MM-PBSA or docking scores of traditional docking software, usually require the atomic structure of the side chains as input, which is not available. Furthermore, they were only validated for discriminating different ligands, while in our design task, we desire to discriminate different binding sites.

\textbf{Baselines.}
\textsc{PiFold} \textit{(no ligand)} is the architecture of \citet{gao2023pifold} and does not use any ligand information and \textsc{ProteinMPNN} \textit{(no ligand)} is the analog with the architecture of \citep{ProteinMPNN}. In \textsc{PiFold} \textit{(2D ligand)}, we first process the ligand with PNA \citep{corso2020principal} message passing and pass its features as additional input to the \textsc{PiFold} architecture. \textsc{Ground Truth Pos} and \textsc{Random Ligand Pos} use the architecture of \textsc{FlowSite} without the ligand structure prediction layers. Instead, the ligand positions are either the ground truth bound structure or sampled from a standard Normal at the pocket's alpha carbon center of mass. {Similarly, \textsc{DiffDock-Pocket Pos} uses fixed positions and the same architecture, but the positions are given by the pocket-level docking tool DiffDock-Pocket \citep{plainer2023diffdockpocket} (we only have results for single ligands since it is not implemented for multi-ligands)}. The oracle \textsc{Ground Truth Pos} method also uses fake ligand data augmentation.

\textbf{Pocket Recovery Results.} Table \ref{table:pocket_recovery} shows that \textsc{FlowSite} consistently is able to recover the original pocket better than simpler treatments of the (multi-)ligand, closing the gap to the oracle method that has access to the ground truth ligand structure. The joint structure generation helps in determining the original residue types (keeping in mind that these are not necessarily the only or best). \textsc{Random Ligand Pos} further confirms that inferring approximate ligand coordinates, like \textsc{HarmonicFlow} in \textsc{FlowSite}, is crucial for recovering the binding pocket. 

\subsection{Question 3: Ablations and Flow Matching}\label{sec:flow_matching_investigation}
\begin{table}
\vspace{-0.4cm}
\caption{\textbf{Flow matching investigation.} Ablations of \textsc{HarmonicFlow}. {Column \%$<$2* indicates performance when selecting the best of 5 generated samples.}}
\label{table:ablations}
\begin{center}
    \begin{tabular}{lccc}
    \toprule
      \rule{0pt}{2ex}
        Method & \%$<$2 & \%$<$2* & Med. \\
    \midrule
    
    {\textsc{Gaussian Prior}}            & {17.0} & {29.2} & {3.8}  \\
    \textsc{Velocity Prediction}        & 11.9 & 28.8 & 3.8  \\
    \textsc{Standard TFN Layers}       & 13.7 & 25.4 & 3.6  \\
    {\textsc{No Refinement Loss}}            & {9.8} & {22.1} & {3.7}  \\
    \textsc{no self-conditioning}       & 14.3 & 29.8 & 3.7  \\
    \midrule
    \textsc{HarmonicFlow} $\sigma=0$                & 18.3 & 31.3 & 3.5  \\
    \textsc{HarmonicFlow} $\sigma=0.5$             & 20.5 & 34.5 & 3.4  \\
    \bottomrule
    \end{tabular}
\end{center}
\vspace{-0.4cm}
\end{table}

Here, we attempt to understand better the behavior of flow-matching generative models for biomolecular structure generation via experiments with \textsc{Harmonic Flow}. We use \textit{Radius-Pockets} on the sequence similarity split of PDBBind. 

\textbf{Investigations.} {\textsc{Gaussian Prior} uses an isotropic Gaussian as prior instead of our harmonic prior.}  In \textsc{Velocity Prediction}, the TFN model predicts $(\vx_1 - \vx_0)$ instead of $\vx_1$ meaning that $\mathcal{L}_{CFM} = \lVert v_\theta - (\vx_1 - \vx_0) \rVert^2$.  In \textsc{Standard TFN layers}, our \hyperref[par:refinement-tfn]{refinement TFN layers} are replaced, meaning that there are no intermediate position updates - only the last layer produces an update. {In \textsc{No Refinement Loss}, the loss $\mathcal{L}_{refine}$ is dropped from the final weighted sum of losses.} \textsc{no self-conditioning} does not use our \hyperref[method:structure-self-conditioning]{structure self-conditioning}. \textsc{Sigma=0} uses $\sigma=0$ for the conditional flow, corresponding to a deterministic interpolant for training.

\textbf{Results.} Table \ref{table:ablations} shows the importance of our self-conditioned flow matching objective, which enables refinement of the binding structure prediction $\Tilde{\vx}^{t}_1$ next to updates of $\vx_t$ at little additional training time - a 12.8\% increase in this experiment. Furthermore, the refinement TFN layers improve structure prediction substantially. Lastly, parameterizing the vector field to predict $\vx_1$ instead of $(\vx_1 - \vx_0)$ appears more suitable for flow matching applications in molecular structure generation.

\section{Conclusion}

We proposed the \textsc{HarmonicFlow} generative process for binding structure generation and \textsc{FlowSite} for binding site design.
Our \textsc{HarmonicFlow} improves upon the state-of-the-art generative process for docking in simplicity, applicability, and performance in various docking settings. We investigated how flow matching contributes to this, together with our technical innovations such as self-conditioned flow matching, harmonic prior ligands, or equivariant refinement TFN layers. 

With \textsc{FlowSite}, we leverage our superior binding structure generative process and extend it to discrete residue types, resulting in a joint discrete-continuous process for designing ligand binding pockets---an important task for which no general and no deep learning solutions exist yet. \textsc{FlowSite} improves upon various baselines in recovering native binding sites without requiring prior knowledge of the bound protein-ligand complex. Thus, \textsc{FlowSite} is a useful step toward generally applicable binding site design, which has important applications in fields ranging from drug discovery to enzyme design.







\clearpage
\section*{Acknowledgments}
We thank Jason Yim, Gabriele Corso, Rachel Wu, Felix Faltings, Jeremy Wohlwend, MinGyu Choi, Juno Nam, Soojung Yang, Nick Polizzi, Jody Mou, Michael Plainer, Julia Balla, Shangyuan Tong, and Peter Holderrieth for insightful discussions and feedback.

This work was supported by the NSF Expeditions
grant (award 1918839: Collaborative Research: Understanding the World Through
Code), the Machine Learning for Pharmaceutical Discovery and Synthesis (MLPDS)
consortium, the Abdul Latif Jameel Clinic for Machine Learning in Health, the DTRA
Discovery of Medical Countermeasures Against New and Emerging (DOMANE)
threats program, the DARPA Accelerated Molecular Discovery program, the NSF AI
Institute CCF-2112665, the NSF Award 2134795, and the GIST-MIT Research Collaboration grant.

\section*{Impact Statement}
We develop a general framework for generating 3D point clouds in \textsc{HarmonicFlow} and for jointly generating discrete and continuous data in \textsc{FlowSite}. As generative modeling frameworks, these are general methods with many possible societal impacts that do not require specific mention. We apply these methods to the tasks of docking and binding site design. In the past and present, the societal impacts of these applications have been overwhelmingly positive, with advances possible in drug discovery, enzyme design, or biological sensing. Future applications with negative outcomes can also be imagined, such as decreasing the barrier of entry for developing biological weapons.
\bibliography{references}
\bibliographystyle{icml2024}

\newpage
\appendix
\onecolumn

\section{Method Details and Explanations} \label{appx:method}
\subsection{BLOSUM Score}\label{appx:blosum_score}
Next to sequence recovery, we also evaluate with our \textit{BLOSUM Score} in an attempt to penalize amino acid predictions less if the predicted residue type is similar yet different from the original residue. With $\mA \in \R^{20 \times 20}$ being the BLOSUM62 matrix, $\mX \in \R^{n \times 20}$ the one hot encoded ground truth residues types and $\hat{\mX} \in \R^{n \times 20}$ the predicted residues types the \textit{BLOSUM Score} is:

\begin{equation}
    Score(\mX,\hat{\mX}) = \frac{\vone^Tdiag(\mX \mA \hat{\mX}^T)}{\vone^Tdiag(\mX \mA \mX^T)}
\end{equation}

\subsection{Fake Ligand Data Augmentation Visualization} \label{appx:fake_ligand_data_augmentation}
\begin{figure}[t]
    \centering
    \includegraphics[width=1.0\textwidth]{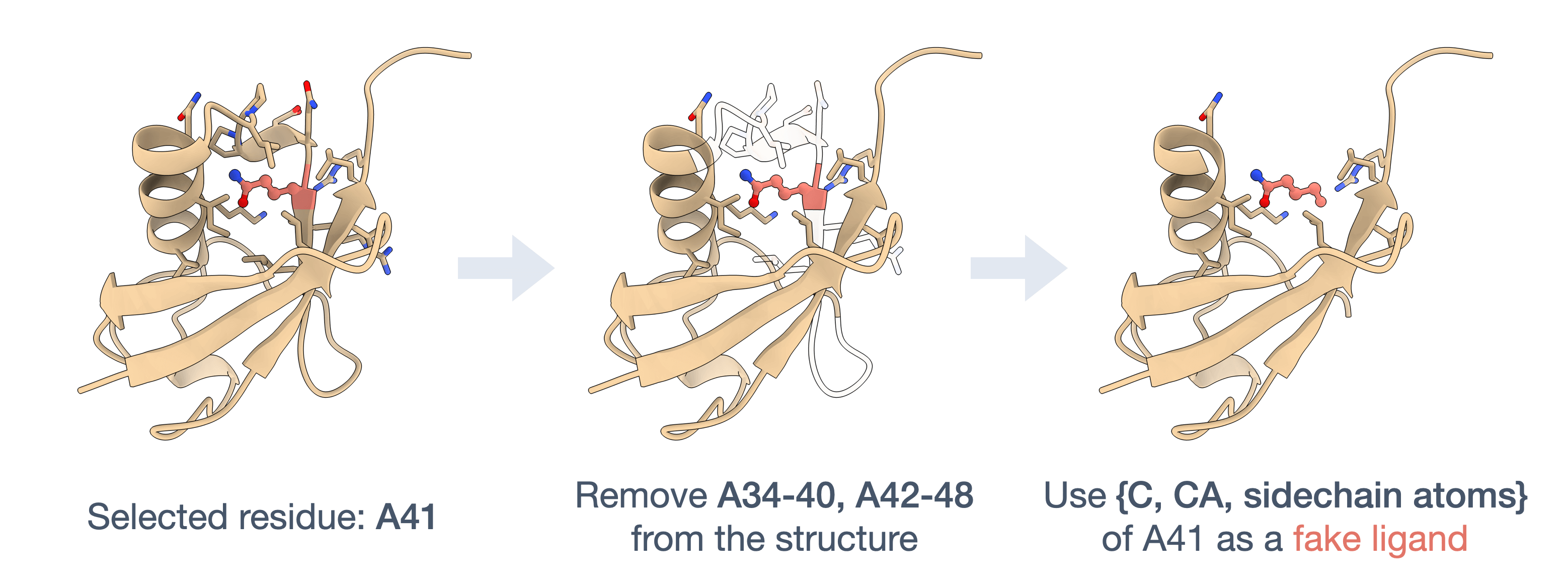}
  \caption{\textbf{Visualization of Fake Ligand creation.} Depicted is a fake ligand created for the Ubiquitin protein. Out of all residues that have at least 4 contacts with other residues (apart from those that are within 7 locations in the chain) a residue is randomly selected as the fake ligand. Then we remove the residue itself from the protein and all residues that are within 7 locations in the chain.} \label{fig:fake_ligand}
\end{figure}

In Figure \ref{fig:fake_ligand}, we visualize the construction of our fake ligands as described in Section \ref{sec:method-fake-ligand}. When constructing the fake ligand from a residue, we drop the backbone oxygen and nitrogen of the amino acid and keep the carbon, alpha carbon, and the side chain as the ligand's atoms.

\subsection{Flow Matching Training and Inference} \label{appx:method_algorithms}

In Section \ref{sec:harmonic_flow}, we lay out the conditional flow matching objective as introduced by \citet{lipman2022flow} and extended to arbitrary start and end distributions by multiple works concurrently \citep{albergo2022building,albergo2023stochastic, pooladian2023multisample, tong2023improving}. We presented conditional flow matching in this more general scenario where the prior $p_0$ and the data $p_1$ can be arbitrary distributions, as long as we can sample from the prior. 

Many choices of conditional flows and conditional vector fields are possible. For different applications and scenarios, some choices perform better than others. We find it to already work well to use a very simple choice of conditional probability path $p_t(\vx|\vx_0,\vx_1) = \mathcal{N}(\vx | t\vx_1 + (1-t)\vx_0, \sigma^2)$, which gives rise to the conditional vector field $u_t(\vx|\vx_0,\vx_1) = \vx_1 - \vx_0$. With this conditional flow and with parameterizing $v_\theta$ to predict $\vx_1$, the optimization and inference is remarkably straightforward as algorithms \ref{alg:training} and \ref{alg:inference} show.
\begin{algorithm}[t]
  \caption{Conditional Flow Matching training with $\vx_1$ prediction and simple constant width gaussian conditional path.}
  \label{alg:training}
  \begin{algorithmic}
\STATE \textbf{Input:} Training data distribution $p_1$, prior $p_0$, $\sigma$, and initialized vector field $v_{\theta}$
\WHILE{Training}
\STATE $\vx_0 \sim p_0(\vx_0)$; $\vx_1 \sim p_1(\vx_1)$; $t \sim \mathcal{U}(0, 1)$
\STATE $\mu_t \gets t\vx_1 + (1-t)\vx_0$
\STATE $\vx \sim \mathcal{N}(\mu_t, \sigma^2 I)$
\STATE $\mathcal{L}_{CFM} \gets \| v_\theta(\vx, t) - \vx_1 \|^2$
\STATE $\theta \gets \mathrm{Update}(\theta, \nabla_\theta \mathcal{L}_{CFM})$ 
\ENDWHILE
\RETURN $v_\theta$ 
\end{algorithmic}
\end{algorithm}

\begin{algorithm}[t]
  \caption{Conditional Flow Matching inference with $\vx_1$ prediction and simple constant width gaussian conditional path.}
  \label{alg:inference}
  \begin{algorithmic}
\STATE \textbf{Input}: Prior $p_0$, number of integration steps T, and trained vector field $v_{\theta}$
\STATE $steps \gets 1$\;
\STATE $\Delta t \gets 1/T$\;
\STATE $t \gets 0$\;
\STATE $\vx_0 \sim p_0(\vx_0)$\;
\STATE $\vx_t \gets \vx_0$\;
\WHILE{$steps \leq T - 1$}
\STATE $\Tilde{\vx}_1 \gets v_\theta(\vx_t, t)$ \;
\STATE     $\vx_t \gets \vx_t + \Delta t(\Tilde{\vx}_1 - \vx_t)/(1-t)$
\STATE $t \gets t + \Delta t$ \;
\ENDWHILE
\RETURN $\vx_t$
\end{algorithmic}
\end{algorithm}

\subsection{Self-conditioned Flow Matching Training and Inference} \label{appx:method_self_conditioned}
In Section \ref{sec:harmonic_flow}, we also explain the self-conditioning training and inference procedure. When additionally using self-conditioning, the training and inference algorithms are only slightly modified and still very simple as presented in algorithms \ref{alg:self_training} and \ref{alg:self_inference}.

\begin{algorithm}[t]
  \caption{Conditional Flow Matching training with $\vx_1$ prediction and simple constant width gaussian conditional path.}
  \label{alg:self_training}
  \begin{algorithmic}
\STATE \textbf{Input:} Training data distribution $p_1$, prior $p_0$, $\sigma$, and initialized vector field $v_{\theta}$
\WHILE{Training}
\STATE $\vx_0 \sim p_0(\vx_0)$; $\vx_1 \sim p(\vx_1)$; $t \sim \mathcal{U}(0, 1)$; $s \sim \mathcal{U}(0, 1)$\;
\STATE     $\mu_t \gets t\vx_1 + (1-t)\vx_0$\;
\STATE     $\vx \sim \mathcal{N}(\mu_t, \sigma^2 I)$\;
\STATE     $\Tilde{\vx}_1 \sim p_0(\Tilde{\vx}_1)$\;
\IF { $ s > 0.5$}
\STATE $\Tilde{\vx}_1 \gets v_\theta(\vx,\Tilde{\vx}_1, t)$)\;
\ENDIF
\STATE $\mathcal{L}_{CFM} \gets \| v_\theta(\vx,\Tilde{\vx}_1, t) - \vx_1 \|^2$\;
\STATE $\theta \gets \mathrm{Update}(\theta, \nabla_\theta \mathcal{L}_{CFM})$ \;
\ENDWHILE
\RETURN $v_\theta$ 
  \end{algorithmic}

\end{algorithm}

\begin{algorithm}[t]
  \caption{Conditional Flow Matching inference with $\vx_1$ prediction and simple constant width gaussian conditional path.}
  \label{alg:self_inference}
  \begin{algorithmic}
\STATE \textbf{Input:} Prior $p_0$, number of integration steps T, and trained vector field $v_{\theta}$
\STATE $steps \gets 1$\;
\STATE $\Delta t \gets 1/T$\;
\STATE $t \gets 0$\;
\STATE $\Tilde{\vx}_1 \sim p(\vx_0)$\;
\STATE $\vx_0 \sim p(\vx_0)$\;
\STATE $\vx_t \gets \vx_0$\;
\WHILE{$steps \leq T - 1$}
\STATE    $\Tilde{\vx}_1 \gets v_\theta(\vx,\Tilde{\vx}_1, t)$ \;
\STATE    $\vx_t \gets \vx_t + \Delta t(\Tilde{\vx}_1 - \vx_t)/(1-t)$ \;
\STATE    $t \gets t + \Delta t$ \;
\ENDWHILE
\RETURN $\vx_t$
\end{algorithmic}
\end{algorithm}

\subsection{\textsc{FlowSite} Architecture} \label{appx:method_architecture}
\begin{figure}[ht]
    \centering
    \includegraphics[width=1.0\textwidth]{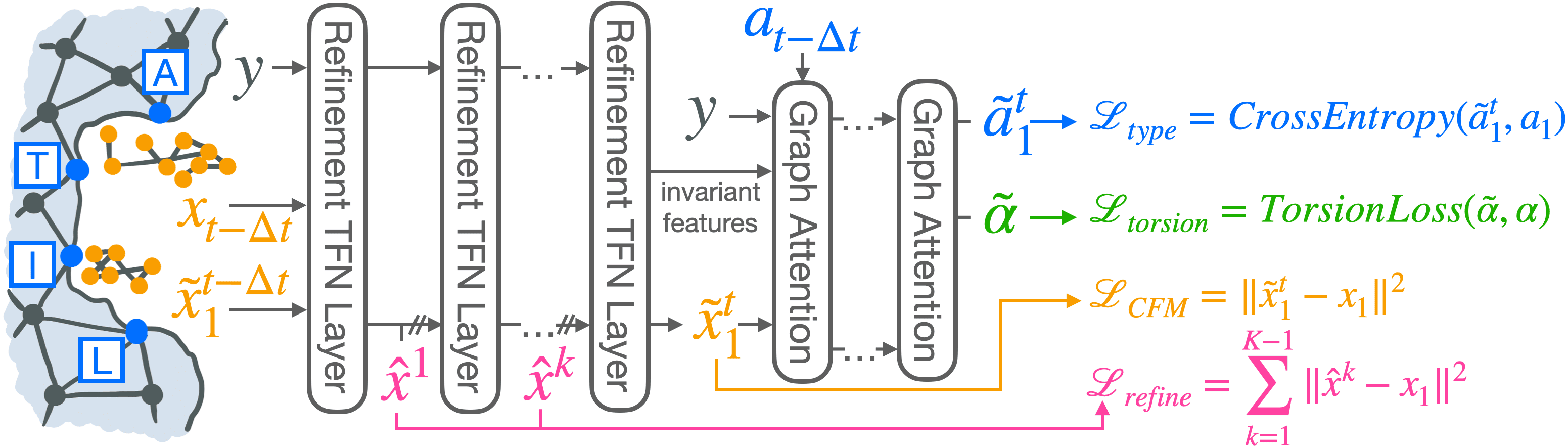}
  \caption{\textbf{\textsc{FlowSite} architecture.} The refinement TFN layers (also in \textsc{HarmonicFlow}) first update the ligand coordinates \textcolor{yellow}{$\vx_{t- \Delta t}$} multiple times to produce the structure prediction \textcolor{yellow}{$\Tilde{\vx}_1^t$} (from which \textcolor{yellow}{$\Tilde{\vx}_{t}$} is computed during inference). The TFN's invariant features and \textcolor{yellow}{$\Tilde{\vx}_1^t$} are fed to invariant layers to produce side chain angles \textcolor{green}{$\Tilde{\mathbf{\alpha}}$} and the new residue estimate \textcolor{strongblue}{$\Tilde{\va}_1^t$}.} \label{fig:architecture}
\end{figure}

Here, we detail the \textsc{FlowSite} architecture as visualized in Figure \ref{fig:architecture} in more detail. The first half of the architecture is an equivariant Tensor Field Network \citep{thomas2018tensor} while the second part is an invariant architecture with graph attention layers similar to the architecture of \textsc{PiFold} \citep{gao2023pifold} where edge features are also initialized and updated.

\textbf{Radius Graph.} The protein and (multi-)ligand are represented as graphs: each residue corresponds to a node, and each ligand atom is a node. Edges are drawn between residue nodes if they are within 50 \AA{}, between ligand nodes if they are within 50 \AA{}, and between the two molecules' nodes if they are within 30 \AA{}. The locations of the residue nodes are given by their alpha carbons, while the atom locations provide the node positions for the ligand nodes. 

\textbf{Node Features.} The ligand features as input to the TNF and to the invariant part of the architecture are atomic number; chirality; degree; formal charge; implicit valence; the number of connected hydrogens; hybridization type; whether or not it is in an aromatic ring; in how many rings it is; and finally, 6 features for whether or not it is in a ring of size 5  or 6. 

The initial receptor features for the TFN are scalar feature encodings of the invariant residue types together with vector features, which are three vectors from the alpha carbon to N, C, and O. 

For the invariant graph attention layer stack, the residue inputs are the invariant geometric encodings of \textsc{PiFold} \citep{gao2023pifold}. Additionally, they contain the residue type self-conditioning information via embeddings of the previously predicted features $\Tilde{\va}_1^{t}$ and the invariant scalar node features of the last refinement TFN layer.

Additionally, radial basis encodings of the sampling time $t$ of the conditional flow are added to all initial node features.

\textbf{Edge Features.} For the Tensor Field Network, the edge features are a radial basis embedding of the alpha carbon distances for the protein-to-protein edges, atom distances for the ligand-to-ligand edges, and alpha carbon to ligand atom distances for the edges between the protein and the ligand. Additionally, the ligand-to-ligand edges features obtain information of the structure self-conditioning by also adding the radial basis interatomic distance embeddings of the previously predicted ligand coordinates $\Tilde{\vx}^t_1$ to them.

Meanwhile, for the invariant graph attention part of the architecture, the ligand-to-ligand edge features are only radial basis embeddings of the interatomic distances. The protein-to-protein edge features are given by radial basis encodings of all pairwise distances between the backbone atoms N, C, Ca, O, and an additional virtual atom (as introduced by \textsc{PiFold}) associated with each residue. The edges between the protein and ligand are featurized as the embeddings of the four possible distances between a single ligand atom and the four backbone atoms of a residue.

\textbf{Tensor Field Network.} The equivariant part of \textsc{FlowSite} uses our equivariant refinement TFN layers based on tensorfield networks \citep{thomas2018tensor} and implemented using the \verb|e3nn| library \citep{e3nn}. These rely on tensor products between invariant and equivariant features. We denote the tensor products as $\otimes_w$ where $w$ are the path weights. Further, we write the $i$-th node features after the $k$-th layer as $\vh^k_i$ for the equivariant Tensorfield network layers. $\vh^0_i$ is initialized as described above in the Node Features paragraph. Lastly, $\mathcal{N}_i$ denotes the neighbors of the $i$-th node in the radius graph.

\textbf{Equivariant TFN Refinement Layer.}
Each layer has a different set of weights for all four types of edges: ligand-to-ligand, protein-to-protein, ligand-to-protein, and protein-to-protein. The layers first update node features before updating ligand coordinates based on them. For every edge in the graph, a message is constructed based on the invariant and equivariant features of the nodes it connects. This is done in an equivariant fashion via tensor products. The tensor product is parameterized by the edge embeddings and the invariant scalar features of nodes that are connected by the edge. To obtain a new node embedding, the messages are summed:
\begin{equation}
\begin{gathered}
{\vh}^{k+1}_i \leftarrow \mathbf{h}^k_i + \textsc{BN} \Bigg( \frac{1}{|\mathcal{N}_i|}\sum_{j \in \mathcal{N}_i} Y(\hat r_{ij}) \; \otimes_{\psi_{ij}} \; \mathbf{h}^k_j \Bigg) \\
\text{with} \; \psi_{ij} = \Psi(e_{ij}, \mathbf{h}^k_i, \mathbf{h}^k_j)
\end{gathered}
\end{equation}
Here, $\textsc{BN}$ is the (equivariant) batch normalization of the \texttt{e3nn} library. The orders of all features are always restricted to a maximum of 1. The neural networks $\Psi$ have separate sets of weights for all 4 kinds of edges. Using these new node features and the previous layer's ligand position update $\hat{\vx}^k$ (or the input positions $\hat{\vx}^0 = \vx_t$ for the first layer), the next ligand position update $\hat{\vx}^{k+1}$ is produced via an O(3) equivariant linear layer $\Phi$ of the \texttt{e3nn} library:

\begin{equation}
\begin{gathered}
\hat{\vx}^{k+1} \leftarrow \hat{\vx}^{k+1} + \Phi({\vh}^{k+1})
\end{gathered}
\end{equation}

\textbf{Invariant Graph Attention Layers.} These layers are based on \textsc{PiFold} and update both node and edge features. The initial features are described in the paragraphs above. We denote these as $\boldsymbol{h}_{i}^{l}$ and $\boldsymbol{e}_{ji}^{l}$ for the $l$-th graph attention layer to disambiguate with the features $\boldsymbol{h}_{i}^{k}$ of the equivariant refinement TFN layers. When aggregating the features for the $i$-th node, attention weights are first created and then used to weight messages from each neighboring node. With $||$ denoting concatenation and $\Omega$, $\Xi$, and $\Pi$ being feed-forward neural networks, the update is defined as:

\begin{equation}
\begin{gathered}
    \label{eq:attention_weight}
        w_{ji} \leftarrow  \Pi(\boldsymbol{h}_j^l || \boldsymbol{e}_{ji}^l || \boldsymbol{h}_i^l)\; \\ 
        a_{ji} \leftarrow  \frac{\exp{w_{ji}}}{\sum_{a \in \mathcal{N}_i}{\exp{w_{ai}}}}\\ \;
       \boldsymbol{v}_j = \Xi(\boldsymbol{e}_{ji}^l || \boldsymbol{h}_j^l)\\ \;
       \boldsymbol{{h}}_i^{l+1} = \sum_{j \in \mathcal{N}_i}{a_{ji} \boldsymbol{v}_j}. \\
   \end{gathered}
\end{equation}

We drop the \textit{global context attention} used in \textsc{PiFold} as we did not find them to be helpful for sequence recovery in any of our experiments. This was with and without ligands.

Based on the new node features, the edge features are updated as follows: 

\begin{equation}
   \label{eq:edge}
      \boldsymbol{e}_{ji}^{l+1} = \Omega(\boldsymbol{{h}}_j^{l+1} || \boldsymbol{e}_{ji}^{l}  || \boldsymbol{{h}}_i^{l+1} ) \\
\end{equation}




\section{Discussion} \label{appx:discussion}
\textsc{HarmonicFlow} has the ability to produce arbitrary bond lengths and bond angles. This distinguishes it from \textsc{DiffDock} \citep{corso2023diffdock}, which only changes torsion angles, translation, and rotation of an initial seed conformer. Thus, unlike \textsc{DiffDock}, \textsc{HarmonicFlow} would be able to produce unrealistic local structures. That this is not the case, as shown in Figure \ref{fig:visualizations} attests to how \textsc{HarmonicFlow} learns physical constraints. Still, we argue that the role of deep learning generative models should be to solve the hard problem of finding the correct coarse structure. If one desires a conformer with low energy with respect to some energy function, this can be easily and quickly obtained by relaxing with that energy function.

\section{Additional Results} \label{appx:results}

\subsection{Multi-ligand Docking Case Studies}
{In this subsection we provide 4 case studies of Multi-liand docking to assess the behaviour of \textsc{HarmonicFlow} for this task. For this purpose, we randomly choose 4 complexes from the Binding MOAD test set under the condition that they contain different ligands.}

{Overall, we find that while the RMSDs of \textsc{EigenFold Diffusion} in Table \ref{table:docking_multi_ligand} are similar to those of \textsc{HarmonicFlow}, the complexes that \textsc{HarmonicFlow} generates are often more physically plausible. For instance, \textsc{HarmonicFlow}'s rings have the appropriate shape and planar systems are actually planar which often is not the case for \textsc{EigenFold Diffusion}.}

\begin{figure}[t]
    \centering
    \includegraphics[width=1.0\textwidth]{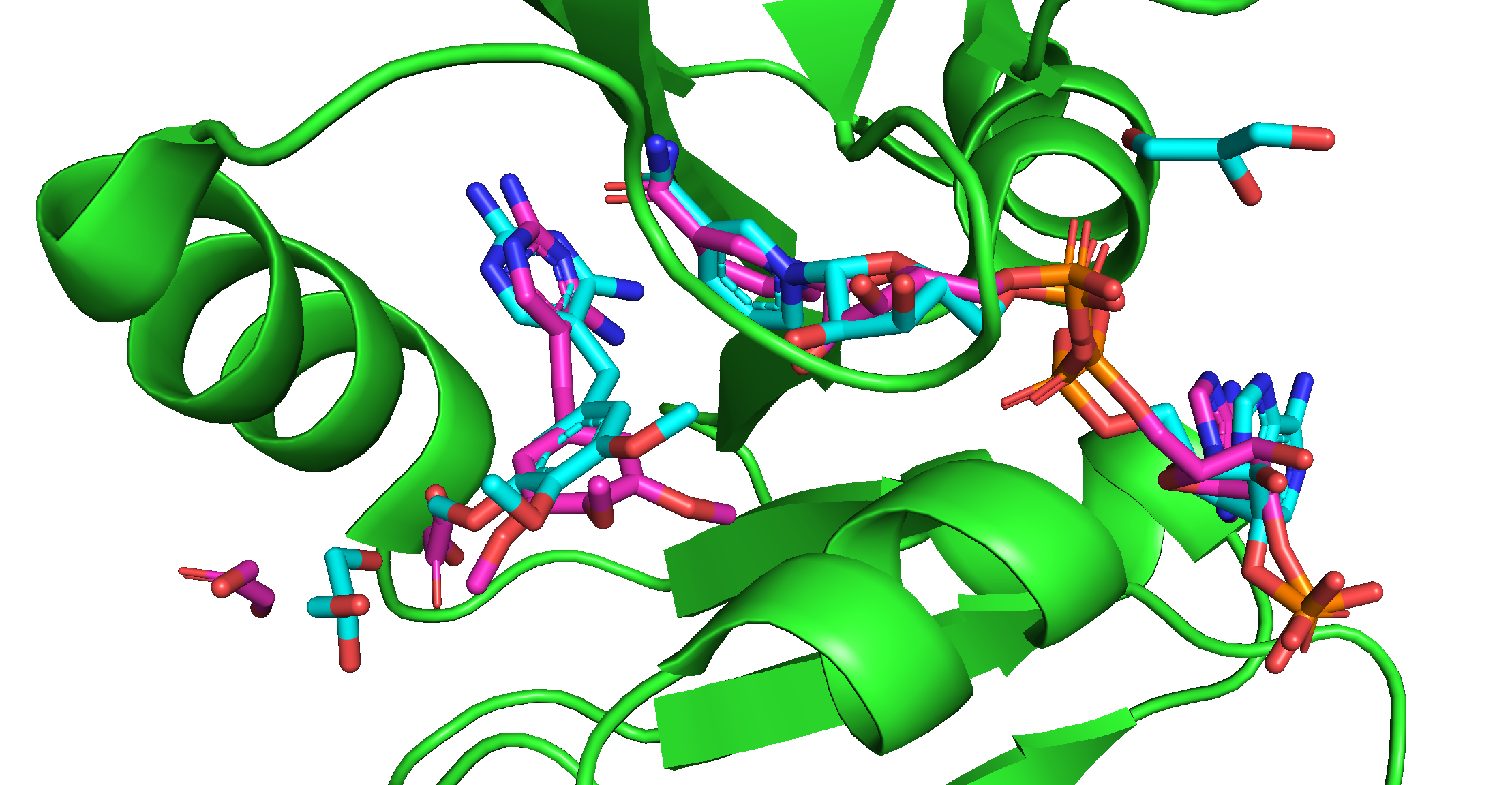}
  \caption{{\textbf{Multi-ligand docking case study.} A randomly picked complex with a multi-ligand from the Binding MOAD test set with 4 molecules as multi-ligand. We show the ground truth ligand structure (blue) and a sample of \textsc{HarmonicFlow} (purple). We find that the predicted structure mostly matches the ground truth for both of the large ligands, but both small ligands are placed on the left while one of them should be on the right.}} \label{fig:case_study_1dg5}
\end{figure}

\begin{figure}[t]
    \centering
    \includegraphics[width=1.0\textwidth]{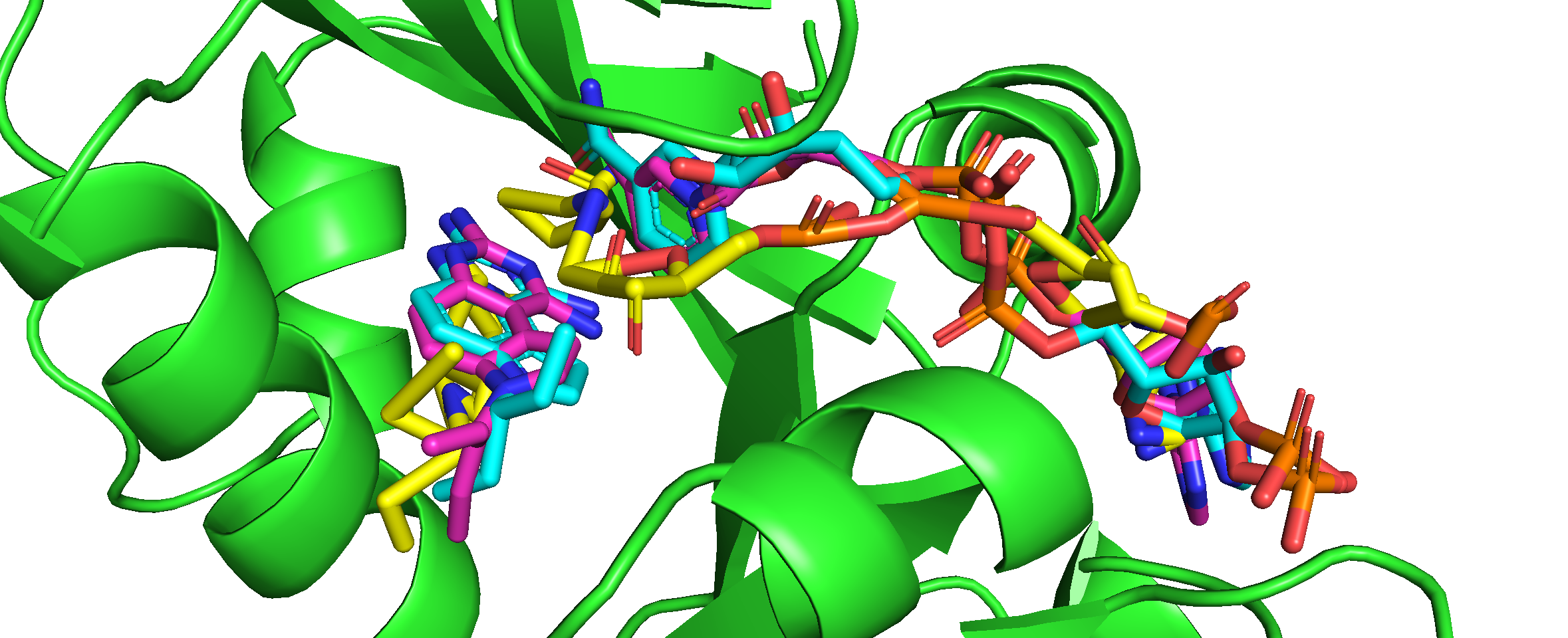}
  \caption{{\textbf{Multi-ligand docking case study.} A randomly picked complex with a multi-ligand from the Binding MOAD test set for which we show the ground truth ligand structure (blue), a sample of \textsc{HarmonicFlow} (purple) and a sample of \textsc{EigenFold Diffusion} (yellow). We find that the rings of \textsc{HarmonicFlow} have the correct shapes and that the bond lengths and angles are physically plausible while \textsc{EigenFold Diffusion} fails to produce planar rings.}} \label{fig:case_study_1aoe}
\end{figure}

\begin{figure}[t]
    \centering
    \includegraphics[width=0.6\textwidth]{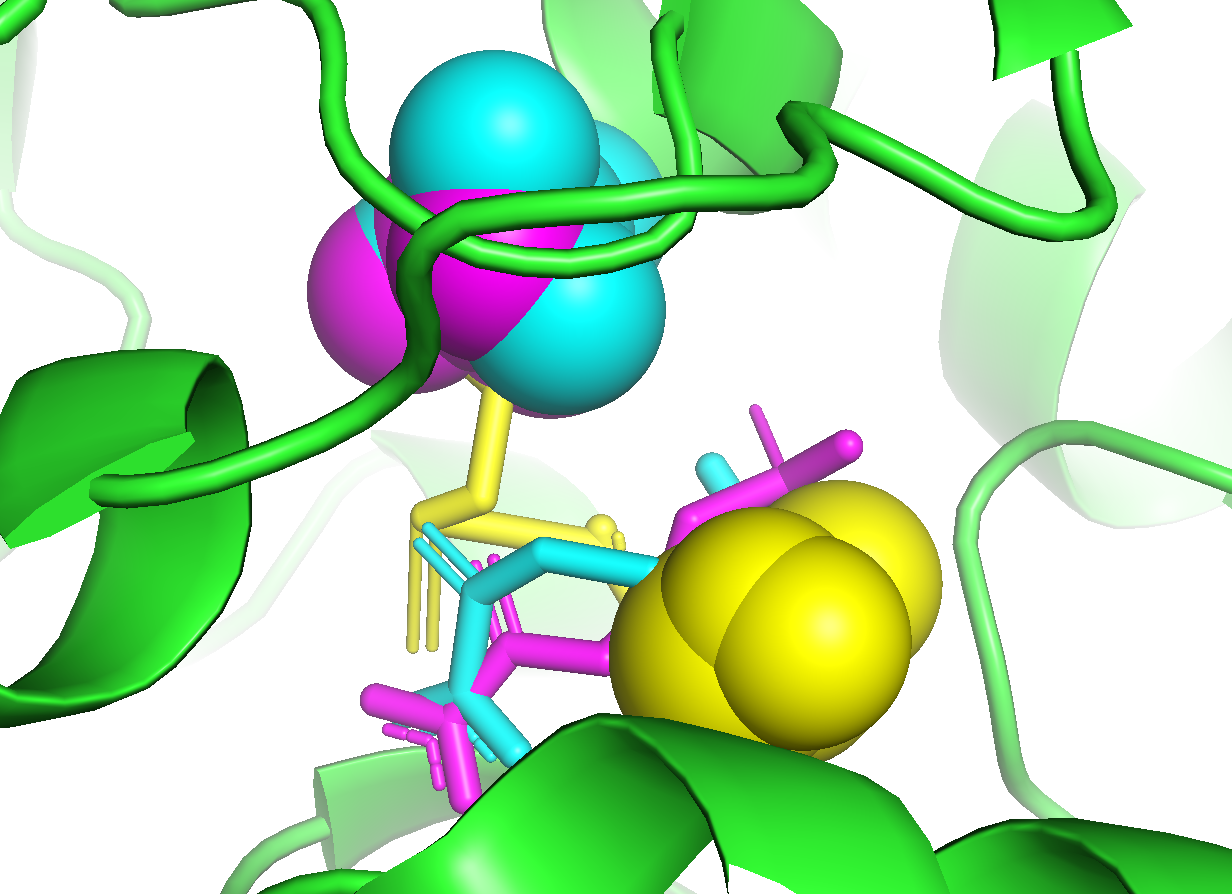}
  \caption{{\textbf{Multi-ligand docking case study.} A randomly picked complex with a multi-ligand from the Binding MOAD test set for which we show the ground truth ligand structure (blue), a sample of \textsc{HarmonicFlow} (purple) and a sample of \textsc{EigenFold Diffusion} (yellow). We find that \textsc{EigenFold Diffusion} incorrectly placed the sulfate and its second ligand prediction is further from the ground truth while \textsc{HarmonicFlow} almost perfectly places the sulfate.}} \label{fig:case_study_1vb3}
\end{figure}

\begin{figure}[t]
    \centering
    \includegraphics[width=0.6\textwidth]{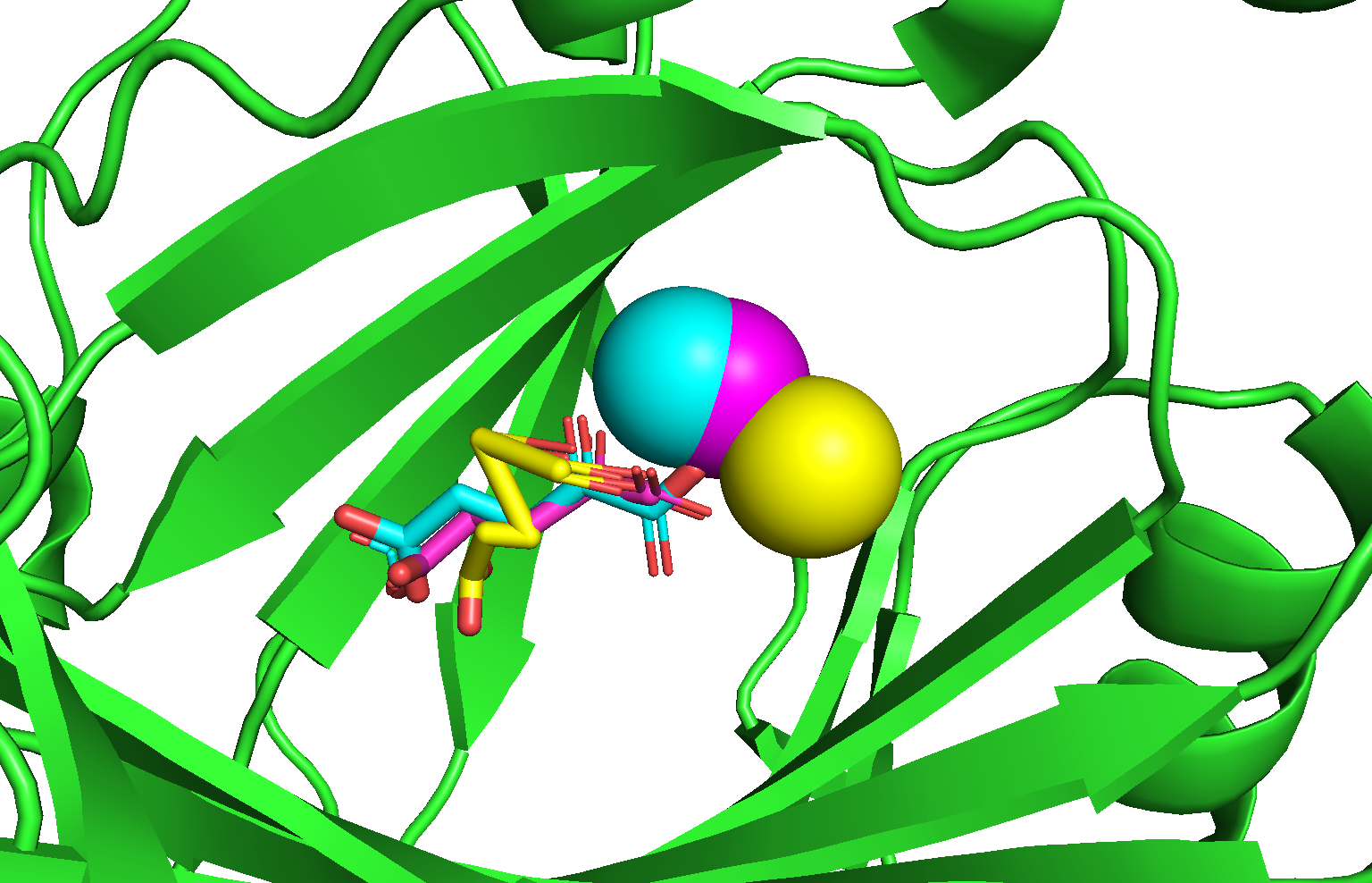}
  \caption{{\textbf{Multi-ligand docking case study.} A randomly picked complex with a multi-ligand from the Binding MOAD test set for which we show the ground truth ligand structure (blue), a sample of \textsc{HarmonicFlow} (purple) and a sample of \textsc{EigenFold Diffusion} (yellow). We can see unrealistic bond lengths and angles for \textsc{EigenFold Diffusion}'s prediction and the predicted ion position is further from the ground truth than for \textsc{HarmonicFlow}.}} \label{fig:case_study_1zmf}
\end{figure}

\subsection{Analysis of joint ODE dynamics}
\begin{figure}
\begin{tabular}{cccc}
\subfloat[PDB ID: 4a80]{\includegraphics[width = 0.4\textwidth]{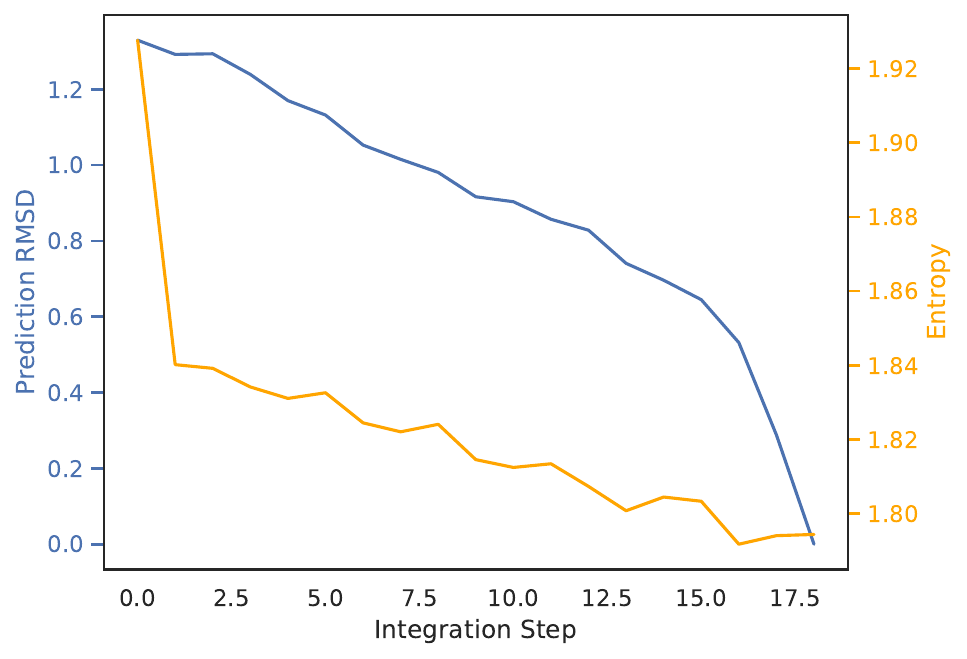}} &
\subfloat[PDB ID: 3hgi]{\includegraphics[width = 0.4\textwidth]{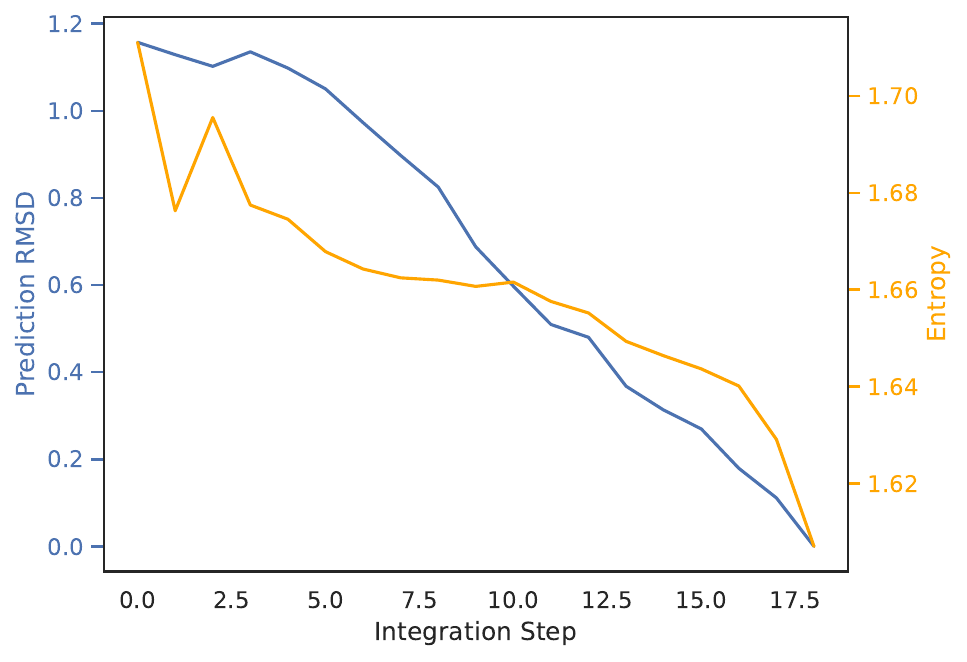}}\\
\subfloat[PDB ID: 6bnd]{\includegraphics[width = 0.4\textwidth]{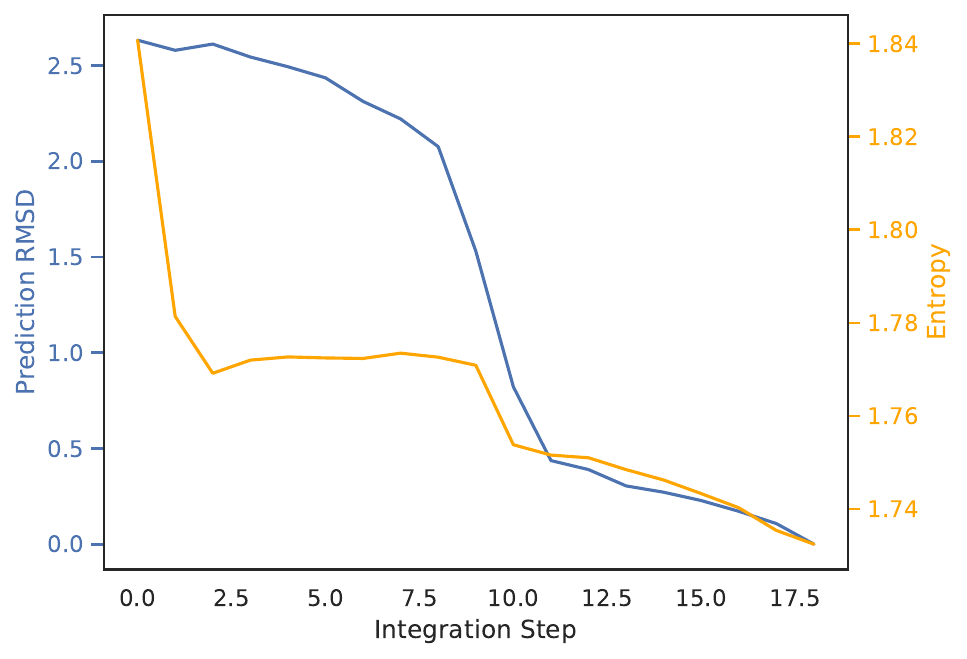}} &
\subfloat[PDB ID: 3i51]{\includegraphics[width = 0.4\textwidth]{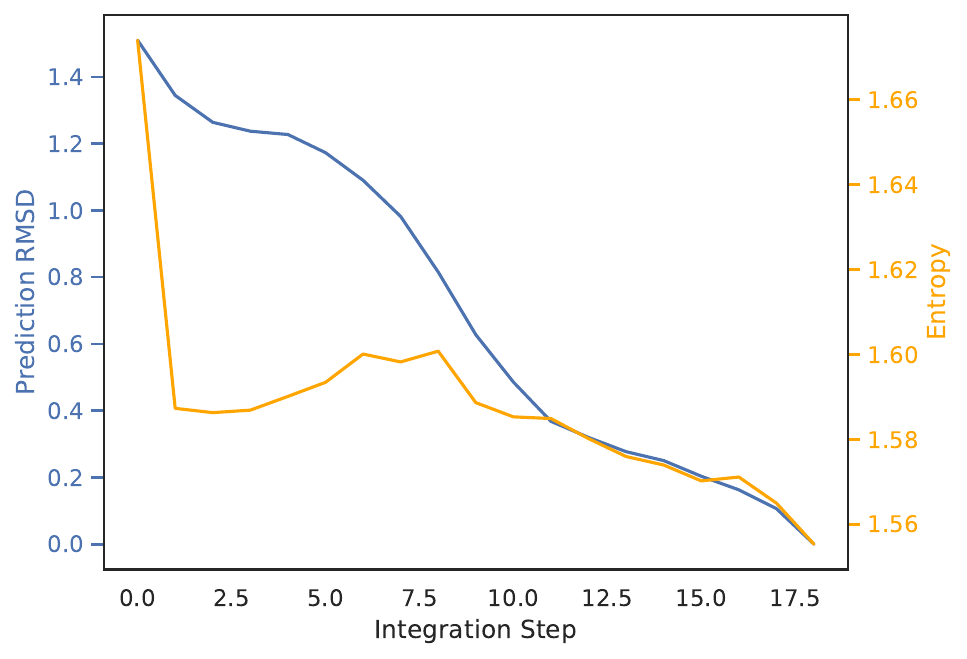}}\\
\subfloat[PDB ID: 2xsu]{\includegraphics[width = 0.4\textwidth]{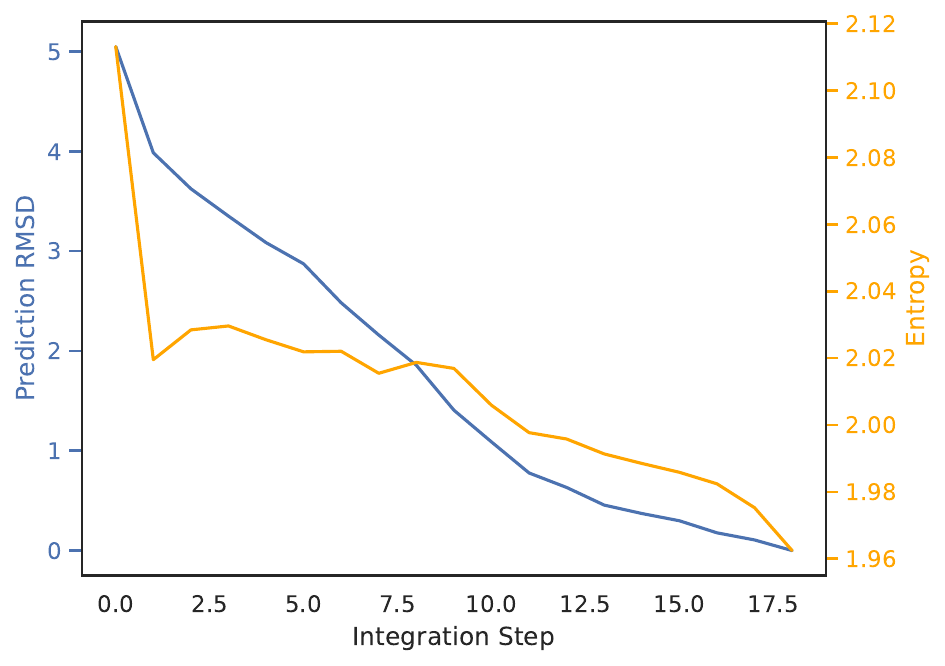}} &
\subfloat[PDB ID: 4p6r]{\includegraphics[width = 0.4\textwidth]{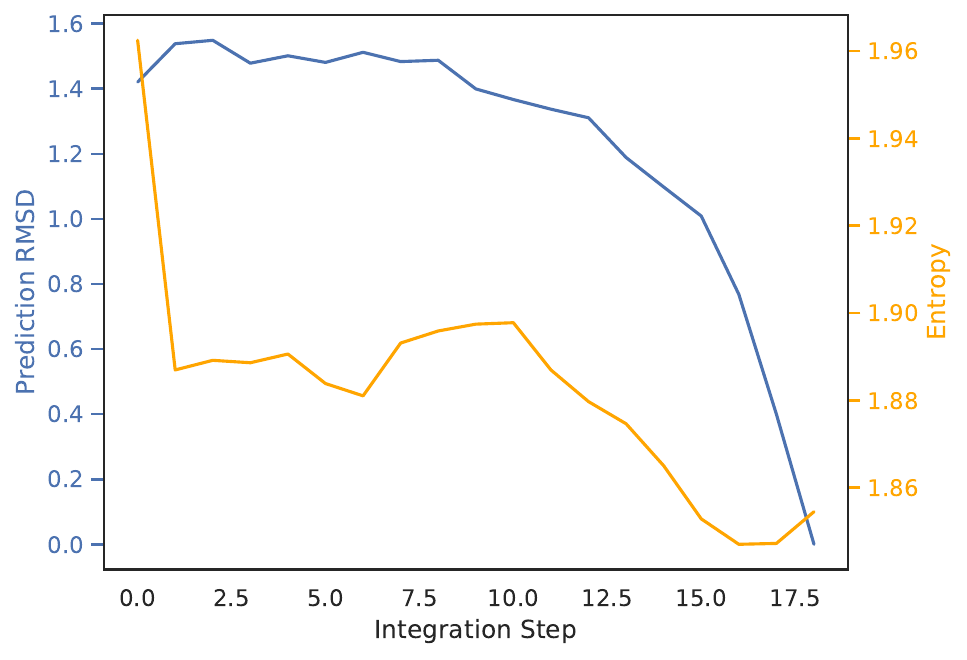}}\\
\subfloat[PDB ID: 1pzp]{\includegraphics[width = 0.4\textwidth]{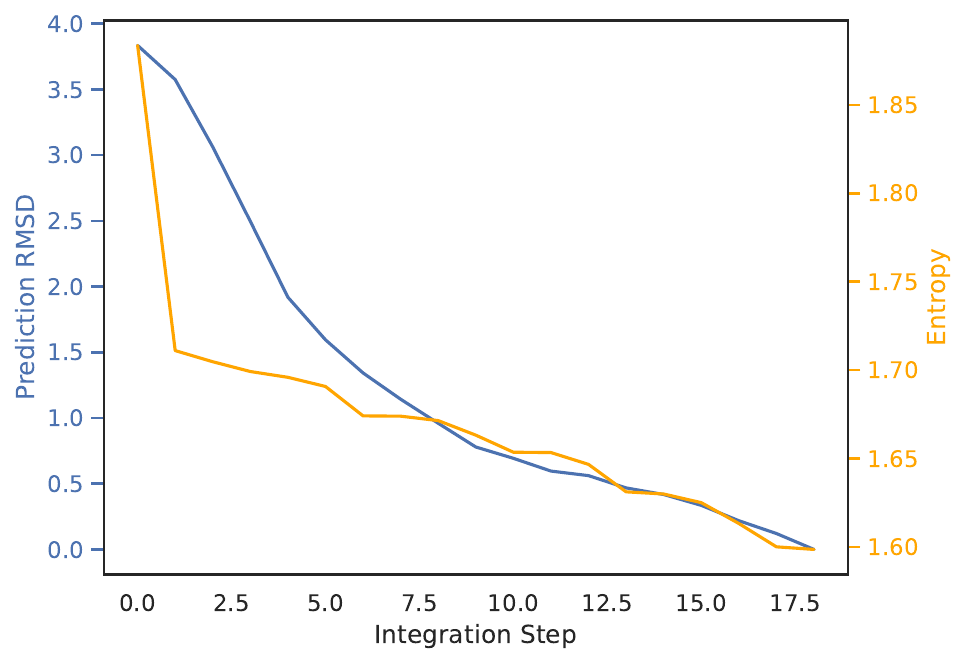}} &
\subfloat[PDB ID: 3hjq]{\includegraphics[width = 0.4\textwidth]{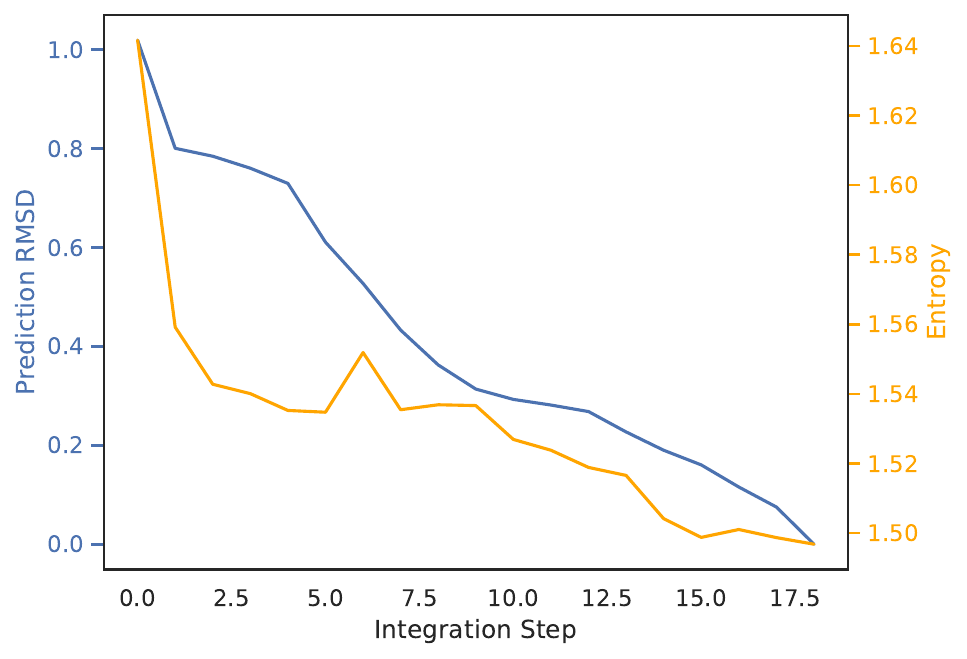}}
\end{tabular}
\caption{{\textbf{Analysis of joint ODE dynamics.} For multiple examples, we visualize two quantities of the trajectory of the ODE. On the x-axis is the integration step of the ODE. On the y-axis on the left in blue, the RMSD between the step's $\vx_1$ prediction and the final $\vx_1$ prediction. On the y-axis on the right in yellow, we show the entropy of the predicted probability distribution averaged over all residues.}} \label{fig:ode_analysis}
\end{figure}

{To analyze the behavior of the learned ODE over discrete and continuous data, we provide Figure \ref{fig:ode_analysis}. This figure shows the evolution of the $\vx_1$ prediction's RMSD and the evolution of the output entropy of the residue type probabilities. The results show how a more determined structure prediction correlates with a decrease in uncertainty of the residue type prediction.} 

\subsection{FlowSite Ablations}
{In Table \ref{table:flowsite_ablation} we provide additional ablations for \textsc{FlowSite}. In \textsc{No Refinement Loss}, the loss $\mathcal{L}_{refine}$ is dropped from the final weighted sum of losses. in \textsc{No Side Chain Torsion Loss}, the loss $\mathcal{L}_{torsion}$ is dropped from the final weighted sum of losses. In \textsc{Backbone Noise}, we add Gaussian noise to the input protein backbone coordinates with standard deviation 0.2. In \textsc{Only Equivariant Layers}, we replace the invariant graph attention layers in the \textsc{FlowSite} architecture with equivariant refinement TFN layers, showing how the invariant architecture is crucial for predicting the discrete residue types.}

\begin{table}[h]
\caption{{\textbf{Binding Site Recovery ablations.} Results on the PDBBind sequence similarity split. \textit{Recovery} is the percentage of correctly predicted residues, and  \textit{BLOSUM score} takes residue similarity into account.}}
\label{table:flowsite_ablation}
\vspace{-0.0cm}
\begin{center}
    \begin{tabular}{lcc}
    \toprule
    \rule{0pt}{2ex}  
        Method &  \textit{BLOSUM score} & \textit{Recovery}    \\
    \midrule
    {\textsc{No Side Chain Torsion Loss}}          & 45.9 & 47.7    \\
    {\textsc{Backbone Noise}}          & 39.6 & 42.4    \\
    {\textsc{No Fake Ligand Augmentation}}          & 45.5 & 46.7   \\
    {\textsc{No Refinement Loss}}          & 45.7 & 47.6    \\
    {\textsc{Only Equivariant Layers}}          & 29.8 & 35.3    \\
    \midrule
    {\textsc{FlowSite} discrete loss weight = 0.8}                   & 46.9 & 47.8     \\
    \textsc{FlowSite} discrete loss weight = 0.2                  & 47.6 & 49.5     \\
    \bottomrule
    \end{tabular}
\vspace{-0.2cm}
\end{center}
\end{table}

\subsection{Docking without residue idenitities}
\begin{table}[ht]
\caption{{\textbf{\textsc{HarmonicFlow} vs. \textsc{Product Diffusion} without residue idenitites.} Comparison on PDBBind splits for docking without residue identities into \textit{Distance-Pockets} (residues close to ligand) and \textit{Radius-Pockets} (residues within a radius of the pocket center). The columns "\%$<$2" show the fraction of predictions with an RMSD to the ground truth that is less than 2\AA{} (higher is better). "Med." is the median RMSD (lower is better).}} \label{table:docking_diffdock_noresidue}
\begin{center}
    \begin{tabular}{lcc|cc|cc|cc}
    \toprule
      & \multicolumn{4}{c}{Sequence Similarity Split} & \multicolumn{4}{c}{Time Split} \\ 
      \rule{0pt}{2ex}
      & \multicolumn{2}{c}{Distance-Pocket} & \multicolumn{2}{c}{Radius-Pocket} & \multicolumn{2}{c}{Distance-Pocket} & \multicolumn{2}{c}{Radius-Pocket} \\
    \rule{0pt}{2ex}  
        Method & \%$<$2 & Med. & \%$<$2 & Med. & \%$<$2 & Med. & \%$<$2 & Med. \\
    \midrule
    \textsc{Product Diffusion}  & 23.2 & 3.4 & 14.3 & 4.0 & 16.9 & 4.3 & 12.3 & 4.6 \\
    \textsc{HarmonicFlow} & 35.5 & 2.8 & 20.8 & 3.5 &  39.3 & 2.8 & 30.7 & 3.3 \\
    \bottomrule
    \end{tabular}
\end{center}
\end{table}
For our binding site design, it is important that the structure modeling of the ligand is accurate given the evidence that having a good model of the (multi-)ligand structure is important for recovering pockets and given the interlink between 3D structure and binding affinity / binding free energy. In the main text Section \ref{sec:experiments_structure_generation}, we investigated \textsc{HarmonicFlow}'s performance for docking with known residue identities. However, when using \textsc{HarmonicFlow} for binding site design, the residue identities are not known a prior, and structure reasoning abilities in this scenario are required.

\subsection{Blind Docking}
\begin{table}[ht]
\caption{{\textbf{\textsc{HarmonicFlow} vs. \textsc{Product Diffusion} for blind docking.} Comparison on PDBBind splits for blind docking where the binding pocket of the protein is not known, and the whole protein is given as input. The columns "\%$<$2" show the fraction of predictions with an RMSD to the ground truth that is less than 2\AA{} (higher is better). "Med." is the median RMSD in \AA{} (lower is better). Top 5 and Top 10 refers to the performance when generating 5 or 10 samples and selecting the best among them.}} \label{table:blind_docking}
\vspace{-0.0cm}
\begin{center}
    \begin{tabular}{lccc|ccc}
    \toprule
      \rule{0pt}{2ex}
      & \multicolumn{3}{c}{Sequence Split} & \multicolumn{3}{c}{Time Split}  \\
    \rule{0pt}{2ex}  
        Method & \%$<$2 & \%$<$5 & Med. & \%$<$2 & \%$<$5 & Med. \\
    \midrule
    \textsc{Product Diffusion}     & 10.7 & 40.6 & 5.9 & 12.6 & 44.1 & 5.6 \\
    \textsc{HarmonicFlow}           & 11.4 & 40.5 & 5.8 & 25.5 & 52.2 & 4.7 \\
    \midrule
                    & Top 5 & Top 10 &  & Top 5 & Top 10 &  \\
                 & \%$<$2 & \%$<$2 &  & \%$<$2 & \%$<$2 &  \\
        \midrule
    \textsc{Product Diffusion}     & 26.6 & 34.2 &  & 28.9 & 32.5 &  \\
    \textsc{HarmonicFlow}           & 17.1 & 20.0 &  & 30.9 & 32.6 &  \\
    \bottomrule
    \end{tabular}
\end{center}
\vspace{-0.3cm}
\end{table}
In blind docking, the binding site/pocket of the protein is unknown, and the task is to predict the binding structure given the whole protein. While in, e.g., drug discovery efforts and in our binding site design task, the pocket is known, many important applications exist where discovering the binding site is necessary. In these experiments, the runs take longer to converge than in the pocket-level experiments. Thus, the DiffDock runs were trained for 500 epochs while the \textsc{HarmonicFlow} runs were trained for 250 epochs instead of the 150 epochs in the pocket-level experiments. Table \ref{table:blind_docking}, shows these preliminary blind docking results, which are promising for a deeper investigation of \textsc{HarmonicFlow} for blind docking and for optimizing it toward this task. This could include training a confidence model as in \textsc{DiffDock} \citep{corso2023diffdock}.

\subsection{Predicted Complex Visualizations}
\begin{figure}[t]
    \centering
    \includegraphics[width=0.9\textwidth]{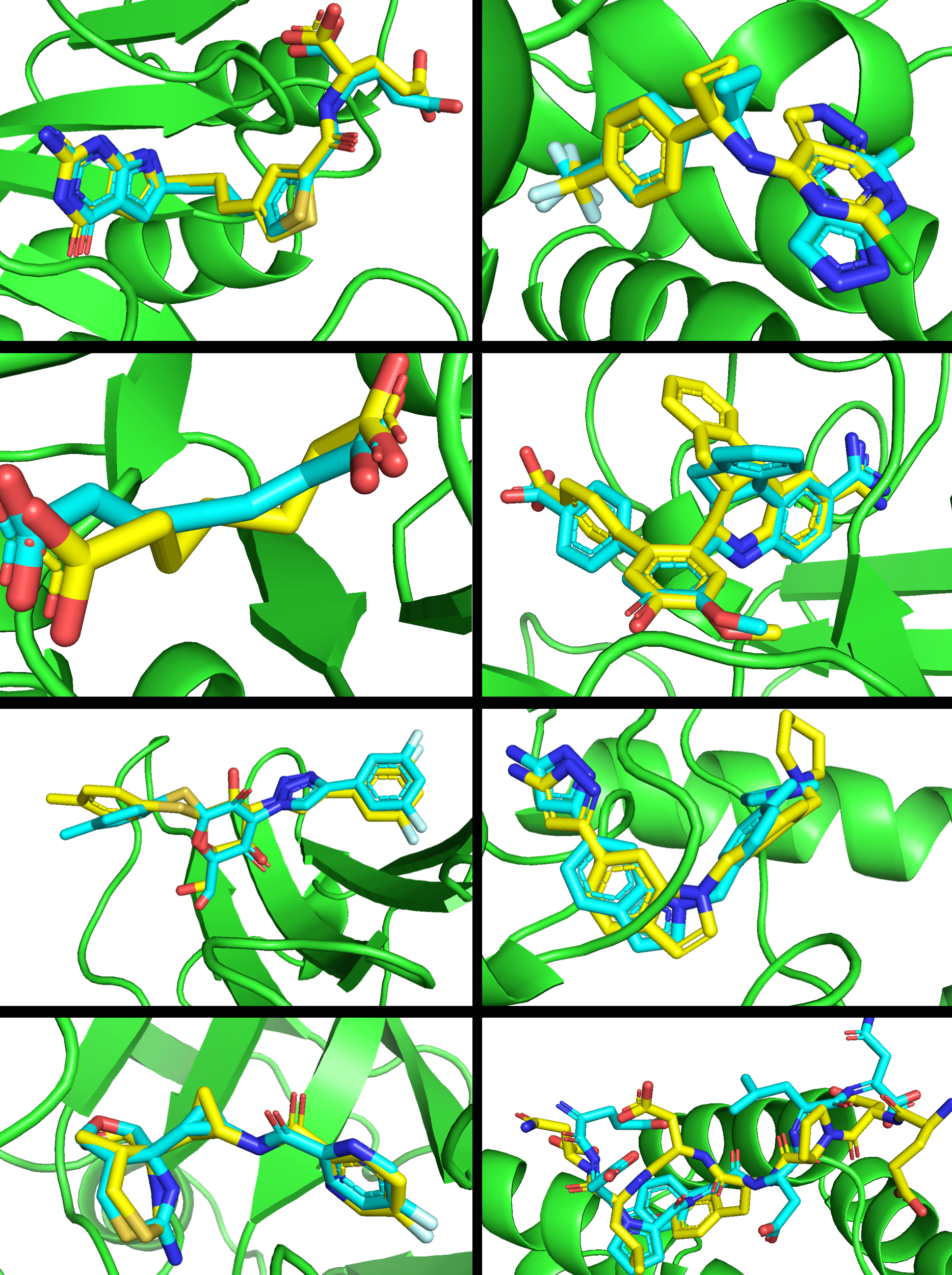}
  \caption{\textbf{\textsc{HarmonicFlow} generated complexes.} Generated complexes of \textsc{HarmonicFlow} for eight randomly chosen complexes in the PDBBind test set in the Distance-Pocket setup with a time-split where none of the ligands were seen during training.} \label{fig:visualizations}
\end{figure}

We visualize generated structures of \textsc{HarmonicFlow} in Figure \ref{fig:visualizations} from the PDBBind test set under the time-based split of \citet{stärk2022equibind} in which there are no ligands whose SMILES string was already in the training data. The generated complexes show very chemically plausible ligand structures even though there are no local structure constraints as in \textsc{DiffDock} and \textsc{HarmonicFlow} has full flexibility in modeling bond angles and bond lengths. 

In Table \ref{table:docking_diffdock_noresidue}, we provide the docking results without residue identities, justifying \textsc{HarmonicFlow}'s use in \textsc{FlowSite} for binding site design.

\section{Additional Related Work} \label{appx:related_wok}
\subsection{Flow Matching, Stochastic Interpolants, and Schrodinger Bridges}
While our exposition of flow matching in the main text focused on the works of \citet{lipman2022flow} and \citet{tong2023improving}, the innovations in this field were made by multiple papers concurrently. Namely, Action Matching \citep{neklyudov2023action}, stochastic interpolants \citep{albergo2022building}, and rectified flow \citep{liu2022flow}  also proposed procedures for learning flows between arbitrary start and end distributions.

An improvement to learning such flows would be if their transport additionally performs the optimal transport between the two distributions with respect to some cost. With shorter paths with respect to the cost \ metric, even fewer integration steps can be performed, and integration errors are smaller. Towards this, \citet{tong2023improving} and \citet{pooladian2023multisample} concurrently propose mini-batch OT where they train with conditional flow matching but define the conditional paths between the optimal transport solution within a minibatch. They show that in the limit of the batch size, the flow will learn the optimal coupling.

This can be extended to learning Schrodinger bridges in a simulation-free manner via an iterative flow-matching and coupling definition procedure \citep{shi2023diffusion} akin to rectified flows. Similarly, \citet{tong2023simulationfree} learn a flow and a score simultaneously to reproduce stochastic dynamics as in a Schrodinger bridge and \citet{somnath2023aligned} learn a Schrodinger bridge between aligned data in a simulation-free manner. Simulation-free here means that the learned vector fields no longer need to be rolled out / simulated during training, which is memory and time-consuming and prohibits learning Schrodinger bridges for larger applications. This was required for previous procedures for learning Schrodinger bridges \citep{debortoli2023diffusion, chen2022likelihood}.

\subsection{Pocket Design}
\citet{Yeh2023} successfully designed a novel Luciferase, which is an enzyme catalyzing the reaction of a Luciferin ligand. In their pipeline, they 1) Choose a protein with a cavity of the right size for the reactants. 2) Keep its backbone 3D structure without amino acid identities and part manually, part computationally decide the amino acids of the cavity to bind the ligands (which we aim to do with \textsc{FlowSite}). 3) Design the rest of the protein's residues with existing tools such as ProteinMPNN \citep{ProteinMPNN}.

The \textsc{PocketOptimizer} line of works \citep{Malisi2012-ad, Stiel2016-bn, Noske2023-th} develops a pipeline for pocket residue design based on physics-inspired energy functions and search algorithms. Starting from a bound protein-ligand complex, \textsc{PocketOptimizer} samples different structures and residue types, which are scored with multiple options of energy functions and optimized with different search algorithms for which multiple options are available. Meanwhile, \textsc{DEPACT} \citep{Chen2022} and \citet{Dou2017sampling} design pockets by searching databases for bound structures of similar ligands, using their protein's residues as proposals, and selecting the best combinations based on scoring functions.

\subsection{Antibody Design}
Another domain where joint sequence and structure design has already been heavily leveraged is antibody design \citep{jin2022iterative, verma2023abode, martinkus2023abdiffuser}. In this task, the goal is to determine the residue types of the complementary determining regions/loops of an antibody to bind an epitope. These epitopes are proteins, and we have the opportunity to leverage evolutionary information. A modeling approach here only has to learn the interactions with the 20 possible amino acids that the epitope is built out of. Meanwhile, in our design task, where we wish to bind arbitrary small molecules, we are faced with a much wider set of possibilities for the ligand.

\subsection{Small molecule design}
Another frontier where designing structure and "2D" information simultaneously has found application is in molecule generation. For instance, \citet{vignac2023digress} and \citet{vignac2023midi} show how a joint diffusion process over a small molecule's positions and its atom types can be used to successfully generate novel realistic molecules. This task was initially tackled by EDM \citep{hoogeboom2022equivariant} and recently was used to benchmark diffusion models with changing numbers of dimensions \citep{campbell2023transdimensional}. 

Often, it is relevant to generate molecules conditioned on context. In particular, a highly valuable application, if it works well enough, would be generating molecules conditioned on a protein pocket to bind to that pocket \citep{lin2022diffbp, schneuing2023structurebased}. These applications would be most prominent in the drug discovery industry, where the first step in many drug design campaigns is often to find a molecule that binds to a particular target protein that is known to be relevant for a disease. In our work with \textsc{FlowSite}, we consider the opposite task where the small molecule is already given, and we instead want to design a pocket to bind this molecule. Here, the applications range from enzyme design (for which the first step of catalysis is binding the reactants \citep{Lehninger}) over antidote design to producing new biomedical marker proteins for use in medicinal diagnosis and biology research.

\subsection{Protein-Ligand Docking}
Historically, docking was performed with search-based methods \citep{trott2010autodock,halgren2004glide,thomsen2006moldock} that have a scoring function and a search algorithm. The search algorithm would start with an initial random conformer and explore the energy landscape defined by the scoring function before returning the best scoring pose as the final prediction. Recently, such scoring functions have been parameterized with machine learning approaches \citep{mcnutt2021gnina, mendez2021deepdock}. In these traditional docking methods, to the best of our knowledge, only extensions of Autodock Vina \citep{trott2010autodock} support multiligand docking. However, this still requires knowledge of the complete sidechains, which is not available in our binding site design scenario.

\section{Experimental Setup Details} \label{appx:experiment_details}
In this section, we provide additional details on how our experiments were run next to the exact commands and code to reproduce the results available at \url{https://github.com/HannesStark/FlowSite}. In all of the paper, we only consider heavy atoms (no hydrogens).

\textbf{Training Details.}
For optimization, we use the Adam optimizer \citep{kingma2014adam} with a learning rate of 0.001 for all experiments. The batch size for pure structure prediction experiments is 4, while that for binding site recovery experiments is 16. To choose the best model out of all training epochs, we run inference every epoch for experiments that do not involve structure modeling and every 5 epochs for the ones that do. The model that is used for the test set is the one with the best metric in terms of sequence recovery or fraction of predictions with an RMSD below 2 \AA{}. When training for binding site recovery, we limit the number of heavy atoms in the ligand to 60. We note that for the structure prediction experiments for Binding MOAD in Table \ref{table:docking_multi_ligand}, the dataset construction for both methods had a mistake where ligands were selected based on their residue ID, which is incorrect because a ligand in a different chain could have the same residue ID - we will correct this in the next version of the manuscript. All models were trained on a single A100 GPU. The models that involve structure prediction were trained for 150 epochs, while those without structure modeling and pure sequence prediction converge much faster in terms of their validation metrics and are only trained for 50 epochs. {On PDBBind, FlowSite took 58.6 hours to train, and on MOAD 115.6 hours, both on an RTX A600 GPU.} The \textsc{DiffDock} models are all trained for 500 epochs. 

\textbf{Runtime.} Averaged over 4350 generated samples, the average runtime of \textsc{HarmonicFlow} is 0.223 seconds per generated structure and that of \textsc{FlowSite} is 0.351 seconds per sequence.

\textbf{Hyperparameters.}
We tuned hyperparameters on small-scale experiments in the Distance-Pocket setup for \textsc{HarmonicFlow} and transferred these parameters to \textsc{FlowSite}, whose additional parameters we tested separately. The tuning for both methods was light, and we mainly stuck with the initial settings that we already found to work well. By default, our conditional probability path  $p_t(\vx|\vx_0,\vx_1) = \mathcal{N}(\vx | t\vx_1 + (1-t)\vx_0, \sigma^2)$ uses $\sigma=0.5$ for which we also tried $0.1, 0.3, 0.5, 0.8$. The number of integration steps we use is 20 for all methods, including \textsc{EigenFold Diffusion} and \textsc{product space diffusion}.  The number of scalar features we use is 32, and we have 8 vector features and 6 of our equivariant refinement TFN layers.

\textbf{\textsc{Product Space Diffusion} baseline.}
This only uses the score model, the diffusion generative model component of \textsc{DiffDock} \citep{corso2023diffdock}. We do not use the confidence model, which is a significant part of their docking pipeline. Such a discriminator could also be used on top of \textsc{HarmonicFlow}, and here, we only aim to compare the generative models. For this, we use the code at \url{https://github.com/gcorso/DiffDock} to train \textsc{Product Space diffusion} with our pocket definitions using the same number of scalar features and vector features using 5 of its default TFN layers followed by its pseudo torque convolution and center-convolution. We train all experiments with \textsc{Product Space Diffusion} for 500 epochs.

\textbf{\textsc{EigenFold Diffusion} baseline.}
Here, we use an identical architecture as for \textsc{HarmonicFlow} and only replace the flow matching training and inference with the diffusion training and inference approach of \textsc{EigenFold} \citep{jing2023eigenfold}. The models were trained in the same settings, and most parameters that we use in \textsc{HarmonicFlow} were first optimized with \textsc{EigenFold Diffusion} since we used it initially.

\section{Dataset Details} \label{appx:data_details}
Here, we lay out the details of the PDBBind and Binding MOAD datasets as we use them to evaluate \textsc{HarmonicFlow}'s docking abilities and \textsc{FlowSite}'s binding site design.

\subsection{PDBBind} \label{appx:data_pdbbind}
\begin{figure}[t]
\begin{center}
\includegraphics[width=.47\textwidth]{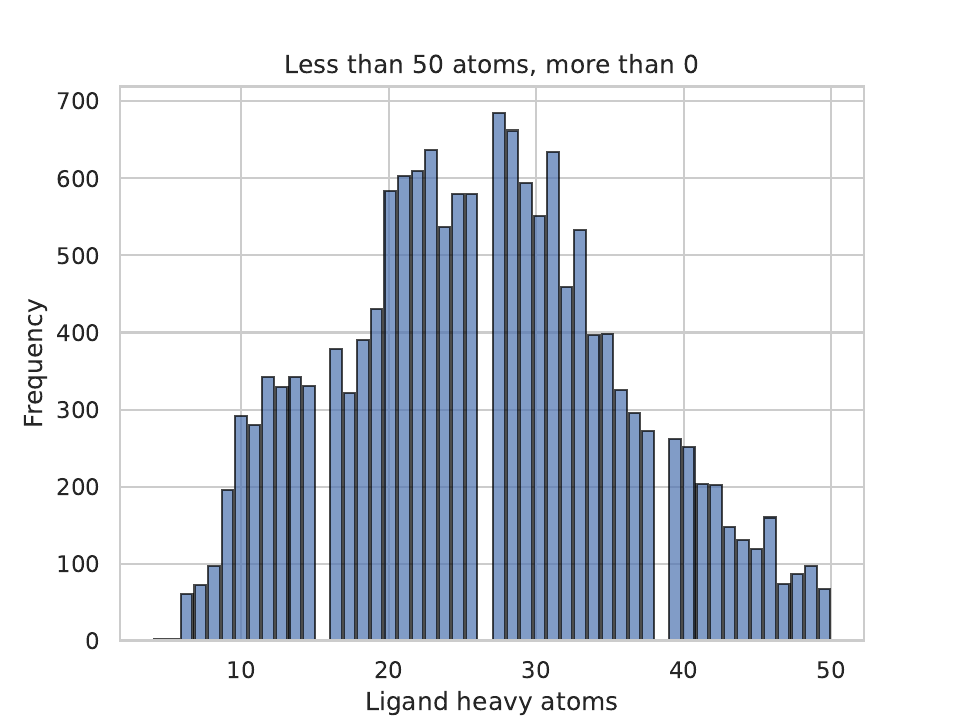}
\includegraphics[width=.47\textwidth]{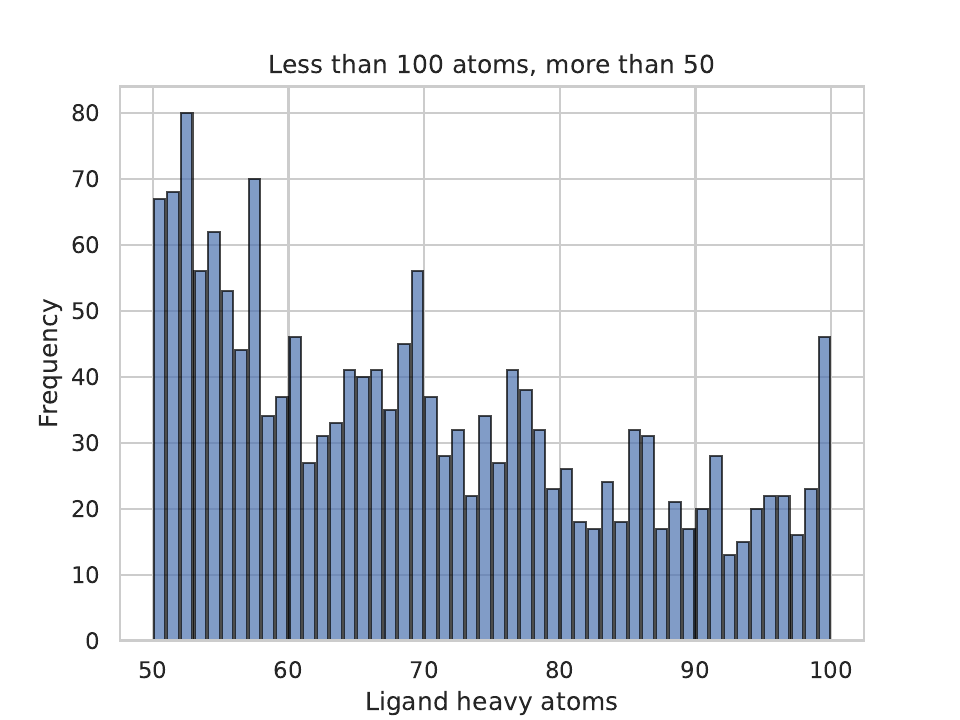}
\caption{\textbf{Number of atoms per ligand: PDBBind.} Histograms showing the number of heavy atoms for all ligands under our ligand definition. This includes many ions, which can be important to filter out if not relevant to the desired application.} 
\label{fig:pdbbind_atoms_per_lig}
\end{center}
\end{figure}

\begin{figure}[t]
\begin{center}
\includegraphics[width=.47\textwidth]{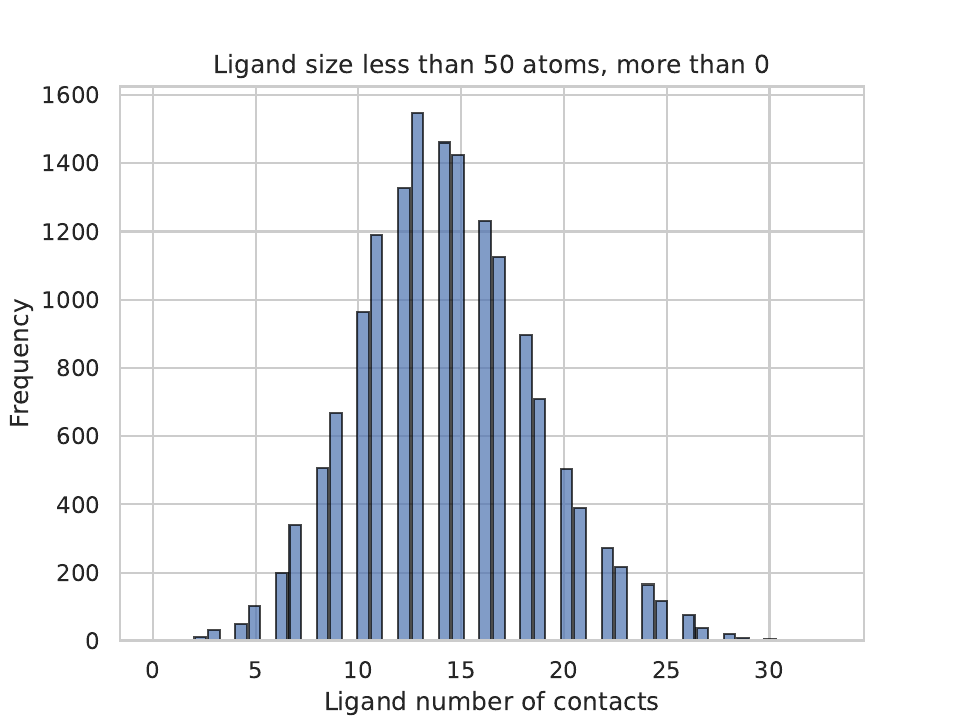}
\includegraphics[width=.47\textwidth]{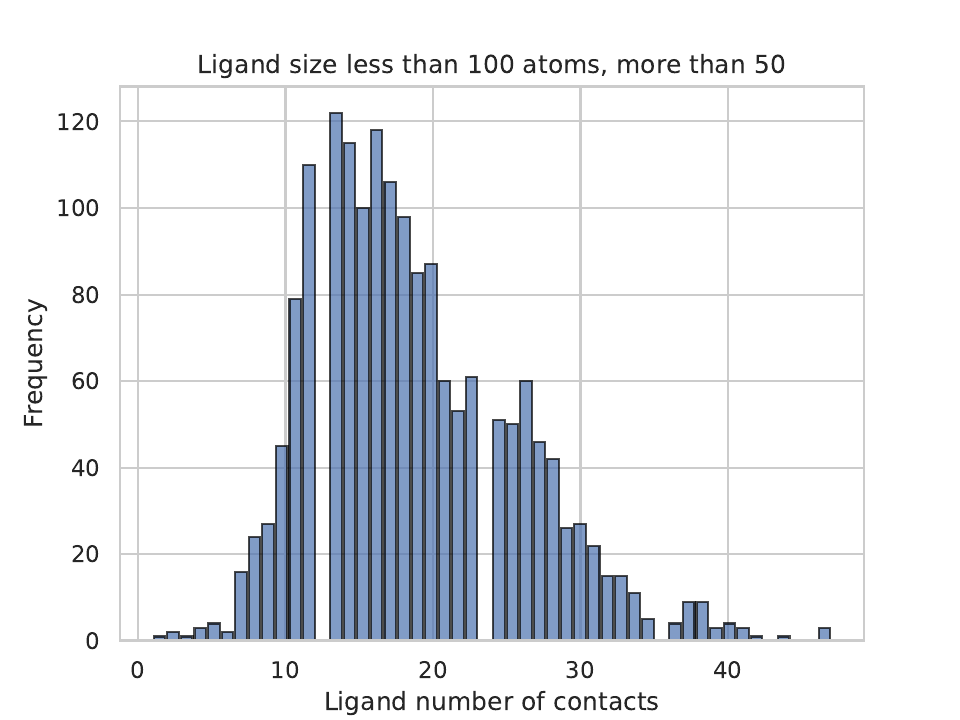}
\caption{\textbf{Number of protein contacts per Ligand: PDBBind.} Histograms showing the number of contacts that each ligand has with its protein. A contact is defined as having a residue with a heavy atom within 4A of any ligand heavy atom.} 
\label{fig:pdbbind_contact_per_lig}
\end{center}
\end{figure}
We use PDBBind dataset \citep{liu2017PDBBind} with protein-ligand complexes of high binding affinity extracted and hand curated from the Protein Data Bank (PDB) \citep{berman2003PDB}. For this, we use two splits. 

\textbf{Splits.}
Firstly, the time split proposed by \citet{stärk2022equibind}, which now is commonly used in the machine learning literature when benchmarking docking approaches, although \citet{buttenschoen2023posebusters} among others found many shortcomings of this split, especially for blind docking. Chiefly among them is the fact that of the 363 test complexes, only 144 are not already included in the training data if a protein is counted the same based on UniProtID. The split has 17k complexes from 2018 or earlier for training/validation, and the mentioned 363 test samples are from 2019. Additionally, there is no ligand overlap with the training complexes based on SMILES identity.  The data can be downloaded from \texttt{https://zenodo.org/record/6408497} as preprocessed by  These files were preprocessed by \citet{stärk2022equibind} with Open Babel before "correcting" hydrogens and flipping histidines with by running \texttt{reduce} \texttt{https://github.com/rlabduke/reduce}. For benchmarking traditional docking software, this preprocessed data should not be employed since the hydrogen bond lengths are incorrect. For our deep learning approaches that only consider heavy atoms, this is not relevant.

Secondly, a sequence similarity, which \citet{buttenschoen2023posebusters} found to be a more difficult split than the time split for the blind docking scenario. To create this split, we cluster each chain of every protein with 30\% sequence similarity. The clusters for training, validation, and test are then chosen such that each protein's chains have at least 30\% sequence similarity with any other chain in another part of the split. This way, we obtain 17741 train, 688 validation, and 469 test complexes. After filtering for complexes that have at least one contact (a protein residue with a heavy atom within 4\AA{}), 17714 train complexes remain while no validation or test complexes are filtered out.

\textbf{Dataset Statistics.}
In Figure \ref{fig:pdbbind_atoms_per_lig}, we show the number of atoms per ligand in two histograms, while Figure \ref{fig:pdbbind_contact_per_lig} shows the number of contacts (a protein residue with a heavy atom within 4\AA{}) per ligand. These statistics are for the training data.

\subsection{Binding MOAD Dataset}
\begin{figure}[t]
\begin{center}
\includegraphics[width=.47\textwidth]{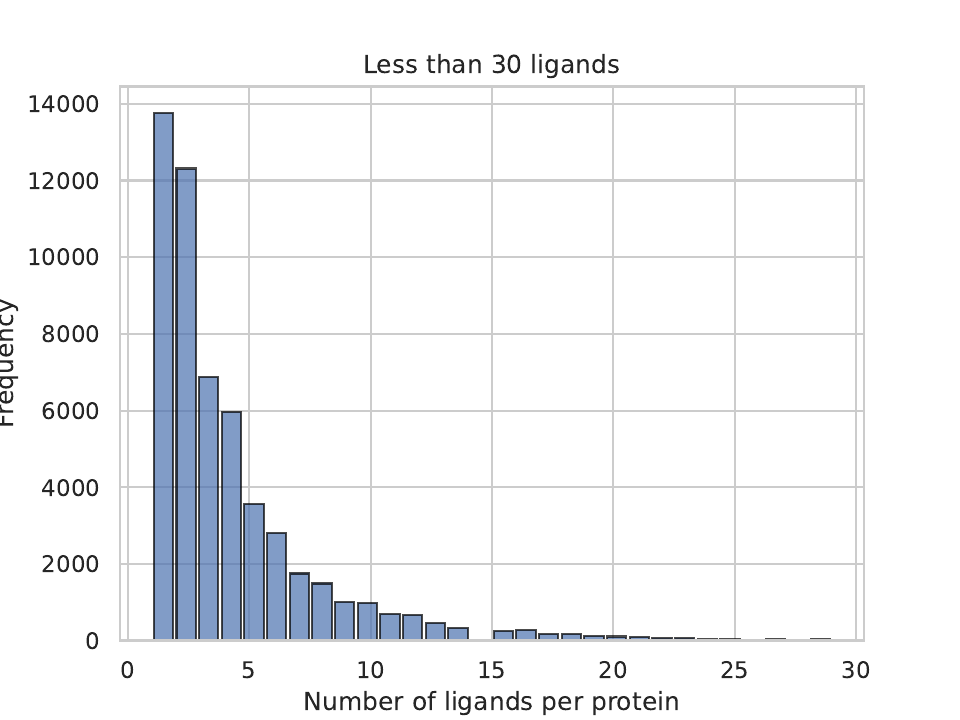}
\includegraphics[width=.47\textwidth]{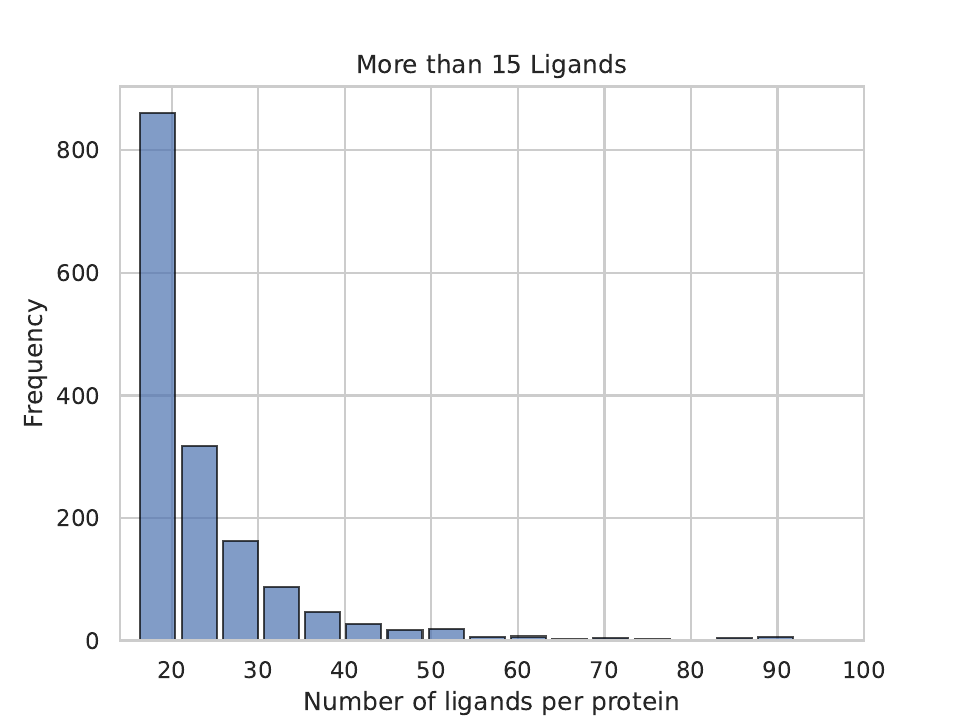}
\caption{\textbf{Number of Ligands per Protein: Binding MOAD.} Histograms showing the number of (multi-)ligands per protein in the Binding MOAD dataset under our ligand definition. Each ligand here can be a multi-ligand. In that case, it is only counted once.} 
\label{fig:moad_lig_per_protein}
\end{center}
\end{figure}

\begin{figure}[t]
\begin{center}
\includegraphics[width=.47\textwidth]{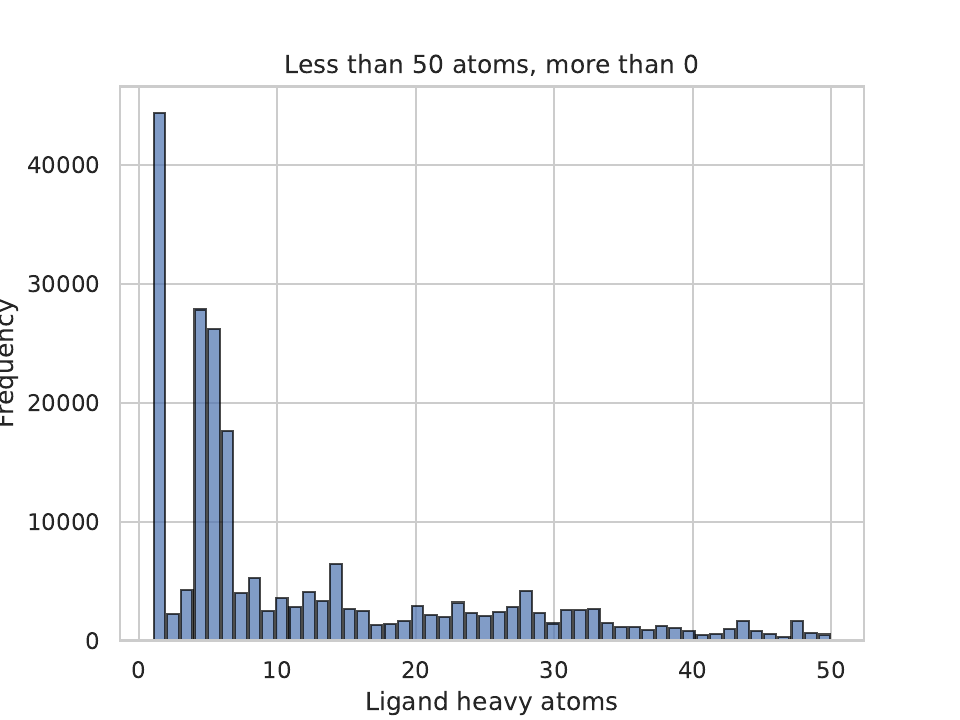}
\includegraphics[width=.47\textwidth]{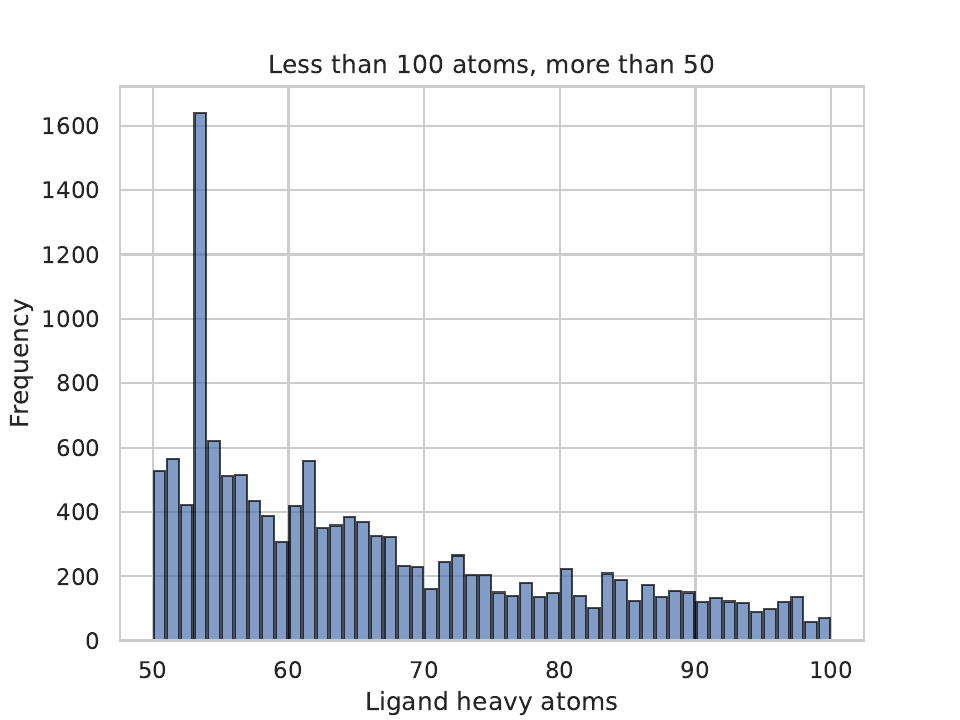}
\caption{\textbf{Number of atoms per ligand: Binding MOAD.} Histograms showing the number of heavy atoms for all ligands under our ligand definition. This includes many ions, which can be important to filter out if not relevant to the desired application.} 
\label{fig:moad_atoms_per_lig}
\end{center}
\end{figure}

\begin{figure}[t]
\begin{center}
\includegraphics[width=.47\textwidth]{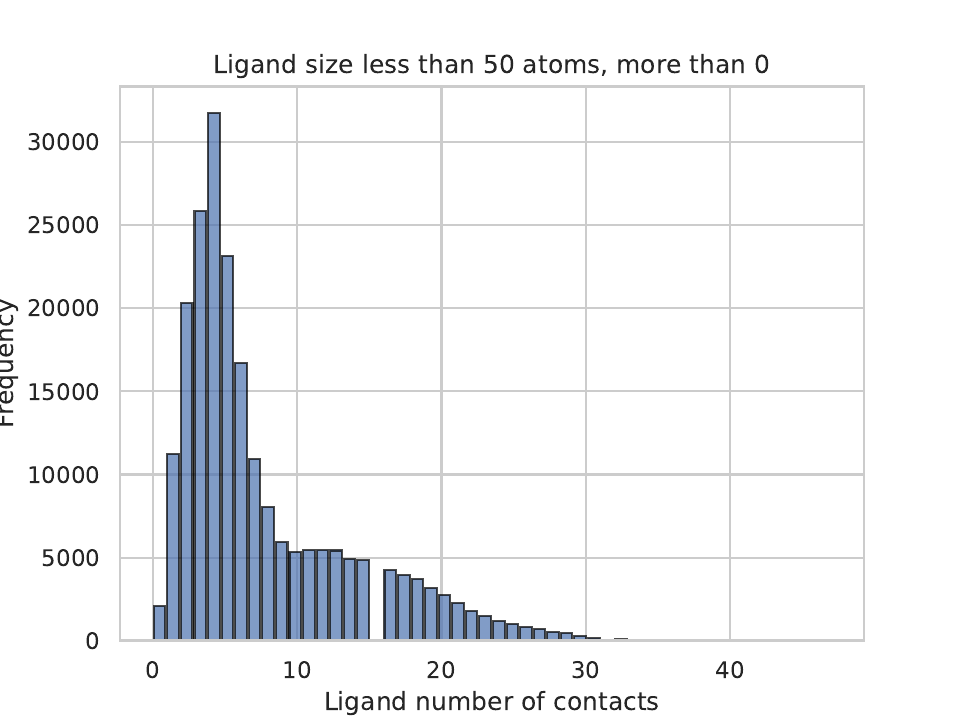}
\includegraphics[width=.47\textwidth]{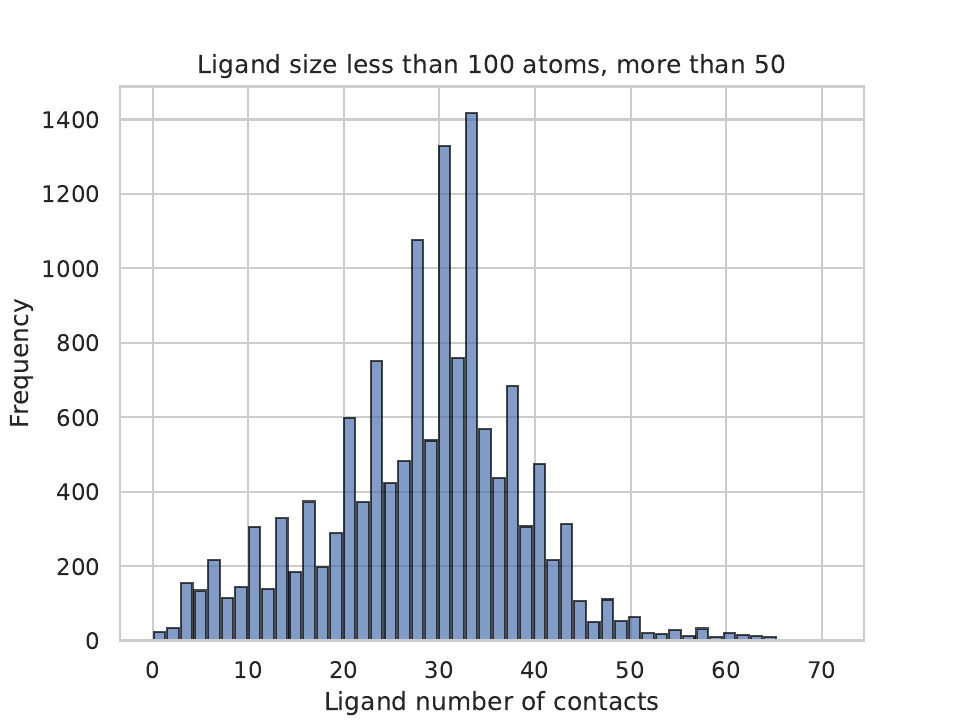}
\caption{\textbf{Number of protein contacts per Ligand: Binding MOAD.} Histograms showing the number of contacts that each ligand has with its protein. A contact is defined as having a residue with a heavy atom within 4A of any ligand-heavy atom.} 
\label{fig:moad_contact_per_lig}
\end{center}
\end{figure}

\textbf{Split.}
The sequence similarity split that we use for BindingMOAD is carried out equivalently as for PDBBind described in Section \ref{appx:data_pdbbind}. This way, we obtain 56649 of Binding MOAD's biounits for training, 1136 for validation, and 1288 as the test set. We discarded some of the biounits and only ended up with 54575 of them since 2.1k of them did not contain any other atoms besides protein atoms and waters. From these, we only use the complexes denoted as the first biounit to reduce redundancy and have only one biounit per PDB ID after which 38477 training complexes remain. We further filter out all ligands that have only one contact (a protein residue with a heavy atom within 4\AA{}) with their protein to obtain 36203 train, 734 validation, and 756 test proteins with a unique PDB ID for each of them.

\textbf{Dataset Statistics.} Here, we provide statistics for the Binding MOAD training data.
In Figure \ref{fig:moad_lig_per_protein}, we show the number of ligands per protein that is obtained under our definition of ligands and multi-ligands. Each ligand in the depicted histogram can either be a multi-ligand or a single molecule. Each multi-ligand is only counted once.
In Figure \ref{fig:moad_atoms_per_lig}, we show the number of atoms per ligand in two histograms, while Figure \ref{fig:moad_contact_per_lig} shows the number of contacts per ligand.

\end{document}